\documentclass[final,10pt,times,onecolumn,numeric]{elsarticle}

\usepackage{amssymb}
\usepackage{color}
\usepackage{xspace}
\usepackage{listings}
\usepackage[ruled, linesnumbered]{algorithm2e}
\usepackage{amsmath}
\usepackage{subcaption}
\usepackage{url}
\usepackage{rotating}
\usepackage{lineno}
\usepackage{float}
\usepackage{hyperref}

\def\ie{i.e.\xspace}
\def\eg{e.g.\xspace}
\def\ourtechnique{\textsc{EvoFlow}\xspace}
\SetKw{KwBy}{by}

\journal{Applied Soft Computing}

\begin{document}

\begin{frontmatter}

\title{Grammar-based evolutionary approach for automated workflow composition with domain-specific operators and ensemble diversity}

\author[uco,dasci]{Rafael Barbudo\corref{x}}\ead{rbarbudo@uco.es}
\author[uco,dasci]{Aurora Ram\'irez\corref{x}}\ead{aramirez@uco.es}
\author[uco,dasci]{Jos\'e Ra\'ul Romero\corref{corauthor}}\ead{jrromero@uco.es}
\cortext[corauthor]{Corresponding author. Tel. +34 957 21 26 60}

\address[uco]{Department of Computer Science and Numerical Analysis, University of C\'ordoba, 14071, C\'ordoba, Spain}
\address[dasci]{Andalusian Research Institute in Data Science and Computational Intelligence (DaSCI), C\'ordoba, Spain \\ \vspace{5px} Published with DOI \href{https://doi.org/10.1016/j.asoc.2024.111292}{10.1016/j.asoc.2024.111292}}

\begin{abstract}
The process of extracting valuable and novel insights from raw data involves a series of complex steps. In the realm of Automated Machine Learning (AutoML), a significant research focus is on automating aspects of this process, specifically tasks like selecting algorithms and optimising their hyper-parameters. A particularly challenging task in AutoML is automatic workflow composition (AWC). AWC aims to identify the most effective sequence of data preprocessing and machine learning algorithms, coupled with their best hyper-parameters, for a specific dataset. However, existing AWC methods are limited in how many and in what ways they can combine algorithms within a workflow.

Addressing this gap, this paper introduces \ourtechnique, a grammar-based evolutionary approach for AWC. \ourtechnique enhances the flexibility in designing workflow structures, empowering practitioners to select algorithms that best fit their specific requirements. \ourtechnique stands out by integrating two innovative features. First, it employs a suite of genetic operators, designed specifically for AWC, to optimise both the structure of workflows and their hyper-parameters. Second, it implements a novel updating mechanism that enriches the variety of predictions made by different workflows. Promoting this diversity helps prevent the algorithm from overfitting. With this aim, \ourtechnique builds an ensemble whose workflows differ in their misclassified instances.

To evaluate \ourtechnique's effectiveness, we carried out empirical validation using a set of classification benchmarks. We begin with an ablation study to demonstrate the enhanced performance attributable to \ourtechnique’s unique components. Then, we compare \ourtechnique with other AWC approaches, encompassing both evolutionary and non-evolutionary techniques. Our findings show that \ourtechnique's specialised genetic operators and updating mechanism substantially outperform current leading methods in predictive performance. Additionally, \ourtechnique is capable of discovering workflow structures that other approaches in the literature have not considered.
\end{abstract}


\end{frontmatter}


\section{Introduction}
\label{sec:intro}

Organisations have accumulated vast amounts of data from diverse sources over the years. However, many of these organisations, particularly small and medium-sized ones, do not analyse these historical data, thus missing valuable insights into their operational activities~\citep{coleman2016}. Extracting useful and novel knowledge from such data is a complex process involving various phases, including problem domain analysis, data integration, dataset preprocessing, model building and deployment, interpretation of results, and decision-making based on study findings~\citep{fayyad1996}.

Although some phases remain inherently human-centric, other phases are prime candidates for (partial) automation~\citep{martinez2019}. A notable example is the automatic selection of  the most appropriate algorithm for model building~\citep{ali2006}. Recently, this idea of automating ML tasks has been formalised in the area of Automated Machine Learning (AutoML)~\citep{barbudo2023}. AutoML provides data scientists with a broader range of alternatives, enabling them to focus on phases that require their expertise and intuition, ultimately bridging the gap between knowledge discovery and domain experts~\citep{feurer2015, chen2018}. Recent studies have even shown that AutoML can outperform humans in specific tasks, such as designing artificial neural network architectures~\citep{zoph2018, chenliang2018}.

Within AutoML, algorithm selection and hyper-parameter optimisation are two of the most commonly addressed tasks~\citep{hutter2019}. Algorithm selection is typically used to predict the optimal model building algorithm~\citep{bilalli2016}, \eg a classifier~\citep{khan2020}. To a lesser extent, it has also been applied to recommend preprocessing algorithms, \eg, the best feature selection algorithm~\citep{parmezan2017}. Hyper-parameter optimisation is predominantly used to fine-tune specific algorithms. However, it should be noted that selecting the best algorithm(s) and potentially optimising their hyper-parameters individually for each phase could impact overall performance. These algorithms are often part of a more complex ML workflow, where algorithms are sequenced and interact through their outputs. Therefore, relationships, synergies, and constraints among them must be comprehensively examined as a whole. To alleviate these shortcomings, some authors have proposed automating workflow composition, involving two or more phases of the knowledge discovery process~\citep{escalante2009,thornton2013}.

Automated workflow composition (AWC) refers to the process of finding the sequence of data processing steps, which typically include data preprocessing, feature selection, and machine learning algorithms, optionally tuning their hyper-parameters, that provides the best performance for a particular machine learning (ML) task (e.g., classification)~\citep{barbudo2023}. AWC represents a challenging task~\citep{heffetz2020} that is commonly approached as an optimisation problem aiming at constructing a workflow that optimally prepares and analyses a dataset. Since a workflow can be specified as an open template in which any combination of preprocessing algorithms is followed by an ML algorithm to build the decision model, the size of the search space can be extremely large. For instance, if we have a high-dimensional classification dataset with missing values, an ideal workflow might start with an imputation algorithm to handle missing data (\eg nearest neighbour imputation) followed by a dimensionality reduction algorithm (\eg PCA), and concluded with a classifier to make predictions (\eg random forest). This workflow is not just a set of algorithms but a sequence where each step is logically and functionally connected to the next, ensuring effective processing of the source dataset. Notice that considering the broad catalogue of algorithms, ways to combine them, and hyper-parameter options, finding the workflow that best fits to input data is time-consuming, lasting several hours or even days.

Proposals for AWC often utilise Bayesian optimisation (BO)~\citep{shahriari2015} and evolutionary algorithms (EAs)~\citep{boussaid2013}. BO is a sequential, model-based optimisation method aimed at reducing the number of evaluations necessary. It does this by selecting the most promising solution(s) for evaluation in each iteration. However, BO-based approaches usually have a predefined structure for the workflows they generate~\citep{feurer2015,thornton2013}, which might restrict their use in certain scenarios. For example, Feurer et al.~\cite{feurer2015} propose optimising workflows that include a feature preprocessor from thirteen options, up to three data preprocessors selected from four alternatives, and a classifier. Similar approaches are seen in recent BO-based proposals, which tend to create even larger workflows~\citep{salvador2018, quemy2020}. In contrast, an EA, an optimisation technique inspired by natural evolution~\citep{boussaid2013}, aims to iteratively enhance a set of solutions (a population) for an optimisation problem. Such an improvement is achieved using variation operators like crossover and mutation, which alter the structure of the individuals, namely their genotype. The objective here is to generate increasingly better adapted individuals, using a fitness function tailored to the problem to evaluate their quality.

EA-based approaches for AWC tend to compose workflows with more flexible structures~\citep{olson2016, larcher2019}, although they still tend to predefine the order and type of preprocessing algorithms applied at each step~\citep{de2017}. Regardless of the technique used, a common practice in the literature is to construct ensembles from the best-performing workflows~\citep{feurer2015}. The purpose is to mitigate overfitting, generally enhancing the generalisation capability of the final ensemble~\cite{bian2021}. Typically, ensembles are built by selecting the top \textit{n} workflows based on predictive performance. However, constructing ensembles solely from workflows with similar predictions may not provide any advantage over using the best workflow, as they could misclassify the same instances. This is especially a concern in evolutionary approaches to AWC, where the optimisation process can lead to a convergence of the population, resulting in workflows that make very similar predictions. Therefore, it seems important to maintain prediction diversity among the selected workflows, offering an advantage over the option of choosing only the best workflow.

In this paper, we introduce \ourtechnique, an approach based on grammar-guide genetic programming (G3P) that enables more flexible and domain-specific workflow structure definitions. G3P is a form of EA in which individuals are represented as tree structures, and their construction is guided by a predefined grammar~\citep{McKay2010}. This grammar ensures the generation of valid individuals that conform to specific syntactic rules, making G3P particularly effective for problems like AWC where the solution structure is complex. For example, the grammar can prevent the creation of workflows where a model building algorithm incorrectly precedes a preprocessing algorithm. The grammar can be adapted to only include interpretable algorithms such as logistic regression or associative classifiers. Specifically, \ourtechnique enhances flexibility and adaptability by allowing workflows to comprise any number of preprocessing steps of any type and order. As we are not imposing restrictions on the resulting algorithm sequence, the solution space is clearly expanded. Notably, G3P has been used by some previous proposals in this domain~\citep{de2017,larcher2019}.

\ourtechnique introduces two novel features that set it apart in the field of AWC. First, unlike other evolutionary proposals that rely on traditional GP operators~\citep{olson2016,de2017}, it implements variation operators like crossover and mutation specifically designed for the AWC problem. These operators consider both the structure and hyper-parameters of workflows, offering a more tailored approach to workflow optimisation. Second, EvoFlow integrates an update mechanism that emphasises diversity in workflow predictions, rather than focusing solely on the best predictive performance. This strategy of promoting prediction diversity is a distinctive innovation compared to other AWC proposals that build ensembles~\citep{feurer2015}, enhancing the robustness and generalisability of the results.
We observed that as optimisation converges, the generated workflows tend to produce highly similar predictions, even when not composed of similar algorithms. To mitigate this limitation, our update mechanism constructs an ensemble considering not only the fitness of the workflows (\ie predictive performance) but also the diversity of their predictions~\citep{bi2012}, thereby mitigating the risk of overfitting. To guide the empirical validation of our proposal, we formulate three research questions (RQs):

\begin{itemize}
    \item RQ1. How do the AWC-specific genetic operators and ensembles contribute to EvoFlow’s model?
    \item RQ2. How does EvoFlow's effectiveness compare to other AWC approaches using different techniques?
    \item RQ3. How does EvoFlow compare in effectiveness to another GP3-based AWC proposal?
\end{itemize}

RQ1 explores the individual and combined effects of EvoFlow’s unique components. RQ2 and RQ3 focus on benchmarking EvoFlow’s performance against other AWC techniques. Our experimental analysis employs 22 classification datasets frequently used to evaluate AWC approaches~\citep{thornton2013, de2017}. The comparison methods are a representative sample of state-of-the-art techniques like BO~\citep{feurer2015}, EAs~\citep{olson2016, de2017}, and automated planning and scheduling (AI planning)~\citep{mohr2018}. Based on this experimentation, the results indicate that the use of our specific genetic operators and ensemble mechanism significantly enhances workflow quality. Furthermore, \ourtechnique significantly outperforms existing approaches in terms of predictive performance. The source code and scripts required to replicate the experiments are publicly available as supplementary material.

The rest of the paper is organised as follows. Section~\ref{sec:background} defines the AWC task, introduces key concepts and terminology related to the applied techniques, and presents the background. Section~\ref{sec:related} reviews related work. Section~\ref{sec:evoflow} provides a detailed description of the proposed method, \ourtechnique. The experimental framework and research questions are described in Section~\ref{sec:experiments}. Then, the experiments are conducted and discussed across three sections: Section~\ref{subsec:experiments_internals} presents an ablation study where the internal components of \ourtechnique are analysed; Section~\ref{subsec:experiments_comparison} compares our proposal against other AutoML approaches; and Section~\ref{sec:experiments_recipe} conducts a specific comparative study against RECIPE (a prior approach applying G3P for AWC). Finally, Section~\ref{sec:conclusions} presents our conclusions and outlines future directions for research.

\section{Background}
\label{sec:background}

This section provides an overview of the most relevant concepts related to the problem automated workflow composition (see Section~\ref{subsec:backgound-awc}) and grammar-based genetic programming (see Section~\ref{subsec:backgound-ec}).

\subsection{Automated Workflow Composition}
\label{subsec:backgound-awc}

AWC involves optimising three related dimensions: (1) algorithms, (2) their relationships, and, optionally, (3) their hyper-parameters values. It is important to note that the algorithm selection problem~\citep{rice1976} specifically refers to recommending the best algorithm(s) for a given dataset and phase of the knowledge discovery process~\citep{barbudo2023}. On the other hand, AWC provides more comprehensive support by covering multiple phases. The relationships between algorithms determine their arrangement within the workflow, allowing them to exploit synergies and handle specific constraints. Finally, the hyper-parameter optimisation problem~\citep{yang2020b} focuses on selecting the optimal values for hyper-parameters associated with algorithms, such as decision tree depth or SVM (Support Vector Machine) kernel. It is worth mentioning that both algorithm selection and hyper-parameter optimisation have been studied independently for decades. More recently, \cite{thornton2013} recognised the need to address both problems jointly, which they referred to as the Combined Algorithm Selection and Hyper-parameter optimisation (CASH) problem. This was approached as a hierarchical optimisation problem, where the selection of one algorithm was considered a hyper-parameter, triggering the optimisation of its respective hyper-parameters while ignoring those of other algorithms. In this paper, we extend the definition of CASH to include the optimisation of relationships between algorithms.

The AWC problem addressed in this paper can be formally defined as follows. Given a set of algorithms $\mathcal{A}$, we define $\mathcal{S}$ as the set of all the possible ordered sequences of these algorithms, ranging in size from 1 to  $|\mathcal{A}|$. It should be noted that algorithms in $\mathcal{A}$ can be applied for data preprocessing ($\mathcal{A_P}$) or model building ($\mathcal{A_{MB}}$), with $\mathcal{A_P} \cap \mathcal{A_{MB}} = \emptyset$. Furthermore, we define $S\in\mathcal{S}$ as a non-empty tuple subject to the following constraints:

\begin{itemize}

\item The size of a sequence $S$, denoted as $|S|$, must satisfy $1 \leq |S| \leq |\mathcal{A_P}| + 1$.

\item For a sequence $S=(a_1, ..., a_i, ... , a_n)$, it must satisfy $a_n \in \mathcal{A_{MB}}$. Moreover,
$\forall i$ such that $1 \leq i < n$, so $a_i \in \mathcal{A_{P}}$.

\end{itemize}

From the above, we observe that any sequence is composed of a model building algorithm (\eg SVM) which may be preceded by one or more preprocessing algorithms (\eg feature selection). Regardless of their type, the performance of algorithms in $\mathcal{A}$ heavily depends on the values of their hyper-parameters. Consequently, the performance of sequence $S$ is determined by the combination of hyper-parameter values of the algorithms it comprises. Let $\lambda$ be the set of hyper-parameters associated with the algorithms in $\mathcal{A}$ and $\lambda_S$ be the hyper-parameters of the algorithms in sequence $S$, such that $\lambda_S \subset \lambda$. Each hyper-parameter $\lambda_1$,$\lambda_2$,...,$\lambda_n$ has its respective domain $\Lambda_1$,$\Lambda_2$,...,$\Lambda_n$, allowing us to define the hyper-parameter space $\Lambda$ as the cross product of these domains, $\Lambda_1\times\Lambda_2\times...\times\Lambda_n$. Furthermore, as a sequence $S$ does not include all the algorithms in $\mathcal{A}$, we define $\Lambda_S$ as the restricted search space that only considers the hyper-parameters of the algorithms in $S$, such that $\Lambda_S \subset \Lambda$. Finally, given a labelled dataset $D$, which is split in $D_{train}$ and $D_{valid}$, the AWC problem aims to find the sequence of algorithms $S^* \in \mathcal{S}$ and their corresponding hyper-parameter values $\Lambda_{S^*} \in \Lambda$ that maximise the predictive performance on $D$. Therefore, the AWC problem can be formally formulated as follows:

\begin{equation}
S^* \Lambda_{S^*} \in \underset{S \in \mathcal{S}, \Lambda_S \in \Lambda}{\mathrm{argmin}} \mathcal{L}(S,\Lambda_S, D_{train}, D_{valid})
\label{ec:awc} 
\end{equation}

where $\mathcal{L}(S_\lambda, D_{train}, D_{valid})$ represents the loss, such as the misclassification rate, obtained by training the sequence $S$ and its corresponding hyper-parameter values $\Lambda_S$ on $D_{train}$ and testing it on $D_{valid}$.

\subsection{Grammar-guided genetic programming}
\label{subsec:backgound-ec}

Evolutionary computation~\citep{back1997} is a field of artificial intelligence that encompasses methods, called evolutionary algorithms, inspired by the evolution of living organisms and are designed to solve complex combinatorial optimisation problems. Different paradigms exist within evolutionary computation, primarily differing in the schema used to represent individuals. Genetic algorithms~\citep{holland1992}, for instance, encode the genotype of individuals as a fixed vector of bits. Another approach is genetic programming (GP), where individuals are encoded as trees without a priori constraints on their shape, size, or structural complexity~\citep{koza1992}. These trees consist of a set of terminal nodes representing operands (\eg 2 or $X$) and internal nodes representing operator functions (\eg + or $\div$). This representation makes GP well-suited for evolving mathematical expressions and computer programs. Notably, GP requires specialised genetic operators to manipulate these tree-based genotypes.

G3P is an extension of GP in which a context-free grammar (CFG) defines the syntactic constraints that must be satisfied by valid individuals. A CFG is defined by a four-tuple $\lbrace S,\sum_N, \sum_T, P\rbrace$, where $S$ is the root symbol, $\sum_N$ is the set of non-terminal symbols, $\sum_T$ denotes the set of terminal symbols, and $P$ defines the set of production rules. It is important to note that the terminal symbols correspond to both operands and operators in GP. A production rule specifies how a non-terminal symbol can be rewritten into one of its derivations until the expression consists only of terminal symbols. Formally, a production rule can be expressed as $a \rightarrow B$, where $a \in \sum_N$ and $B \in \lbrace\sum_N \cup \sum_T\rbrace^*$. In G3P, each individual is created by deriving a unique sequence of production rules, represented as a derivation tree. The elements of the CFG are taken into account during the application of crossover and mutation operators to ensure the generation of valid individuals.

\section{Review Work}
\label{sec:related}

A pioneering work in the field of AWC was proposed by~\cite{escalante2009}, who used particle swarm optimisation to address the complete model selection problem. This involves selecting the best algorithms, together with their hyper-parameters, to perform feature scaling, feature selection, and classification tasks on a given dataset. Subsequently, \cite{thornton2013} formalised the CASH problem, which was tackled by Auto-WEKA. This tool, a BO-based approach, automatically composes a two-step workflow consisting of a feature selection algorithm and a classifier, both taken from the WEKA software\footnote{Weka 3: https://www.cs.waikato.ac.nz/ml/weka/ (last accessed: January 19, 2024)}. Auto-WEKA has been extended to include regression algorithms~\citep{kotthoff2017} and other types of preprocessing algorithms~\citep{salvador2018}. HyperOpt-Sklearn~\citep{komer2014} and Auto-Sklearn~\citep{feurer2015} are two closely related BO-based approaches that use algorithms taken from scikit-learn\footnote{scikit-learn: https://scikit-learn.org/ (last accessed: January 19, 2024)} to construct their workflows. However, the latter also applies meta-learning to warm-start BO and combines prediction from the most accurate workflows found in an ensemble. This approach has been further enhanced by incorporating a new model selection strategy, a portfolio-building mechanism, and an automated policy selection method~\citep{feurer2020}. Given that evaluating a single workflow can be time-consuming (it may take hours or even days), several BO-based proposals have introduced mechanisms to accelerate the process, such as surrogate models~\citep{nguyen2021}, caching algorithms~\citep{zhang2016}, and parallelism with Apache Spark~\citep{anderson2017}, among others. 

As highlighted by Quemy~\citep{quemy2020}, most AWC approaches are based on a fixed sequence of generally few steps with a strict ordering, which limits their applicability to specific scenarios or domains. However, evolutionary algorithms tend to generate more complex workflows. A notable example is TPOT~\citep{olson2016}, which employs GP to construct multi-branch classification workflows. The optimisation process is guided by a two-objective fitness function aiming at maximising the classification accuracy and minimising the workflow size. On the other hand, evolutionary approaches require the evaluation, \ie training, of a larger number of workflows, which can be computationally prohibitive for large datasets. To address this, TPOT has been extended to train workflows on a subset of the data and use the entire dataset only for the most promising ones~\citep{gijsbers2017, parmentier2019}. The use of tree structures allows more complex workflows to be generated~\citep{kvren2017}, as they are not limited to fixed-size genotypes like other evolutionary paradigms, \eg evolution strategy~\citep{burger2015}. Moreover, some researchers have employed grammars to define the structure of a valid workflow, enabling them to avoid, for example, the application of a neural network to a dataset with categorical features. Both G3P~\citep{de2017, larcher2019} and grammatical evolution~\citep{estevez2020, assunccao2020} have been applied to guide the optimisation process. However, their grammars still impose restrictions on the workflow structure. Specifically, RECIPE~\citep{de2017} enforces the order in which the different types of preprocessing algorithms can be executed. AutoML-DSGE~\citep{assunccao2020} adapts the RECIPE grammar by defining a grammar for each dataset. Also, Auto-CVE~\citep{larcher2019} composes workflows with an optional data preprocessing algorithm, an optional feature selection algorithm, and a classifier. Similarly, HML-Opt~\citep{estevez2020} constructs workflows with the same three steps, where each step is recursively composed of different algorithms within that category. All these approaches employ genetic operators derived from the GP literature.

The AWC problem has been addressed using various optimisation techniques or combinations thereof. Notably, AI planning techniques, which were proposed before the definition of the CASH problem, have been applied. In this field, \cite{kietz2012} and \cite{katz2020} use hierarchical task network (HTN) planning to compose workflows based on an ontology and a CFG, respectively. Similarly, \cite{mohr2018} proposed ML-Plan, which, in addition to HTN planning, incorporates a specially designed mechanism to prevent overfitting. Although to a lesser extent, other techniques applied for AWC are reinforcement learning~\citep{elkholy2019, heffetz2020}, multi-armed bandit~\citep{das2018}, and particle swarm optimisation~\citep{escalante2009, diaz2018}, among others. It is important to note that some approaches separate the selection of algorithms and their relationships from the setting of their hyper-parameters. Therefore, they first apply Monte-Carlo tree search~\citep{rakotoarison2019}, meta-learning and multi-armed bandit~\citep{shang2019}, or BO~\citep{quemy2020} to make algorithm selection, before adjusting hyper-parameters with BO.

As mentioned earlier, \ourtechnique is a G3P-based technique for AWC that, unlike most existing approaches in the literature~\citep{li2023}, does not enforce a prefixed workflow structure. More precisely, compared to current grammar-based approaches, it does not restrict the type of preprocessing algorithms to be applied or their order within the workflow sequence. This enables \ourtechnique to expand the solution space and promote greater diversity among solutions. Moreover, domain-specific genetic operators dedicated to optimising both the structure and hyper-parameters of workflows have been developed for \ourtechnique. Lastly, in contrast to current proposals that focus on ensembles based solely on the predictive performance of the workflow~\citep{feurer2015, burger2015}, \ourtechnique also considers the workflow prediction diversity. This avoids the construction of ensembles composed of workflows that lead to the same or similar predictions, even when they are composed of different algorithms and/or different hyper-parameters.

\section{\ourtechnique}
\label{sec:evoflow}

The general procedure of \ourtechnique is outlined in Algorithm~\ref{alg:evoflow}. This algorithm takes the following inputs: the maximum number of generations (\textit{maxGen}), the size of the population (\textit{popSize}), the grammar (\textit{cfg}), the maximum number of derivation steps (\textit{maxDer}), the probabilities for crossover (\textit{cxProb}) and structural mutation (\textit{stMutProb}), the number of individuals to be returned (\textit{archSize}), the training set (\textit{train}), the maximum allowed time for the optimisation process in seconds (\textit{budget}), the allowed time for evaluation (\textit{evalBudget}), and the weight assigned to diversity (\textit{divWeight}), which can range from 0 to~1. The output of \ourtechnique is an external archive (\textit{archive}) that includes the most accurate and diverse individuals, \ie the workflows. 

\begin{algorithm}[H]
    \small
    \SetKwInOut{Input}{In}
    \SetKwInOut{Output}{Out}
    \SetKwProg{try}{try}{:}{}
    \SetKwProg{catch}{catch}{:}{end}
    \Input{maxGen, popSize, cfg, maxDer, cxProb, stMutProb, archSize, train, budget, evalBudget, divWeight}
    \Output{archive}
    \textit{\% Handling initial population} \\
    pop $\leftarrow$ genWorkflows(popSize, cfg, maxDer) \\
    \try{}{
      evaluate(pop, train, evalBudget) \\
      archive $\leftarrow$ update(pop, archSize, divWeight) \\
      
      \While{gen $<$ maxGen}{
        \textit{\% Selecting a pair of parents} \\
        pop $\leftarrow$ select(pop, popSize) \\
        \textit{\% Randomly applying a crossover operator} \\
        \For{$i\gets0$ \KwTo popSize \KwBy $2$}{
          \If{random() $<$ cxProb}{
            \eIf{commonHparams(pop[i], pop[i+1]) $\geq$ 2}{
                cxHparams(pop[i],~pop[i+1]) \\
            } {
          	   cxStruct(pop[i],~pop[i+1]) \\
            }
          }
        }
        \textit{\% Randomly applying a mutation operator} \\
        \For{$i\gets0$ \KwTo popSize}{
  	      \eIf{random() $<$ stMutProb}{
       	    mutStruct(pop[i],~cfg,~maxDer) \\
       	  } {
       		mutHparams(pop[i], cfg) \\
      	  }
        }
        \textit{\% Evaluating offspring and updating the archive} \\
        evaluate(pop, train, evalBudget) \\
        archive $\leftarrow$ update(pop $\cup$ archive, archSize, divWeight) \\
        gen++
         }
         \Return archive
    }
    \vspace{0,2cm}
    \catch{TimeoutException(budget)}{
        \textit{\% Exception thrown: Timed out reached} \\
        \Return archive
     } 
    \caption{\ourtechnique}
    \label{alg:evoflow}
\end{algorithm}

\begin{equation}
fitness = \frac{1}{|classes|} \sum_{c \in classes} \frac{TP_c}{TP_c + FN_c}
\label{ec:bacc}    
\end{equation}

Regarding its operation, the algorithm initially generates a random population, denoted as \textit{pop}, consisting of \textit{popSize} individuals. These individuals are created in accordance with the specified grammar \textit{cfg} (line 2). Notice that the \textit{maxDer} parameter limits the number of times non-terminal symbols can be derived using production rules, effectively constraining the size of the derivation tree and the number of algorithms in the workflows. 
The individuals in \textit{pop} are then evaluated, with their fitness being calculated within the allocated time frame of \textit{evalBudget} seconds (line 4). In this paper, we use balanced accuracy as the fitness function, given that our experimentation focuses on classification datasets and this metric is commonly adopted by other proposals in the field. Balanced accuracy, which should be maximised, is calculated by determining the recall of each class, followed by computing the average across all classes (see Equation~\ref{ec:bacc}). Any individuals exceeding this time budget are assigned a fitness value of 0. To mitigate the risk of overfitting, a 5-fold cross-validation is implemented on the \textit{train} dataset. This number of folds was chosen based on preliminary experiments that showed its effectiveness in striking a balance between predictive generalisation and evaluation time. This approach is in line with standard practices in the field, as exemplified by its default use in MLPlan~\cite{mohr2018}. The external archive \textit{archive} is initially populated with the \textit{archSize} individuals demonstrating the highest fitness (line 5). The algorithm proceeds through its iterations until it reaches either the \textit{maxGen}-th generation (line 6) or the \textit{budget} time limit (line 33).

During each generation, the selection operator picks \textit{popSize} individuals (parents) from the current \textit{pop} population (line 8). A crossover operator, either \textit{cxHparams} or \textit{cxStruct}, is applied to each pair of parents with a probability of \textit{cxProb} (lines 11--18), meaning not all parent pairs undergo recombination. The choice of genetic operators and their specifics will be elaborated upon in Section~\ref{subsec:evoflow_operators}. For now, it is important to note that \textit{cxHparams} is only applicable to parents with at least two hyper-parameters in common (line 12), while \textit{cxStruct} is used in other cases (line 15). Similarly, two mutation operators, \textit{mutStruct} and \textit{mutHparams}, are applied based on the \textit{stMutProb} probability (lines 20--26), with \textit{mutHparams} being selected only when \textit{mutStruct} is not. In practice, a low \textit{stMutProb} prioritises \textit{mutHparams}, allowing hyper-parameter values to be modified without changing the workflow structure, unlike \textit{mutStruct}. Finally, the external archive is updated to retain individuals with the best fitness while ensuring diversity in their predictions (line 29). The importance of fitness and diversity is determined by the \textit{divWeight} parameter.

Upon completing \textit{maxGen} generations or exceeding the \textit{budget} time limit, the \textit{archive}, containing up to \textit{archSize} workflows, is returned. These workflows, both accurate and diverse, are then employed to form an ensemble using a weighted majority voting scheme. The weight of each workflow is computed by dividing its fitness by the fitness of the best individual. It is important to note that after the evolutionary algorithm concludes, these individuals are retrained on the complete \textit{train} set, as the cross-validation is performed during evaluation. Detailed explanations of individual enconding, genetic operators, and the archive update procedure will be provided in the subsequent sections.

\subsection{Encoding}
\label{subsec:evoflow_encoding}

\begin{figure}[H]
\centering

\begin{subfigure}{\linewidth}
    \centering
    \includegraphics[width=0.7\linewidth]{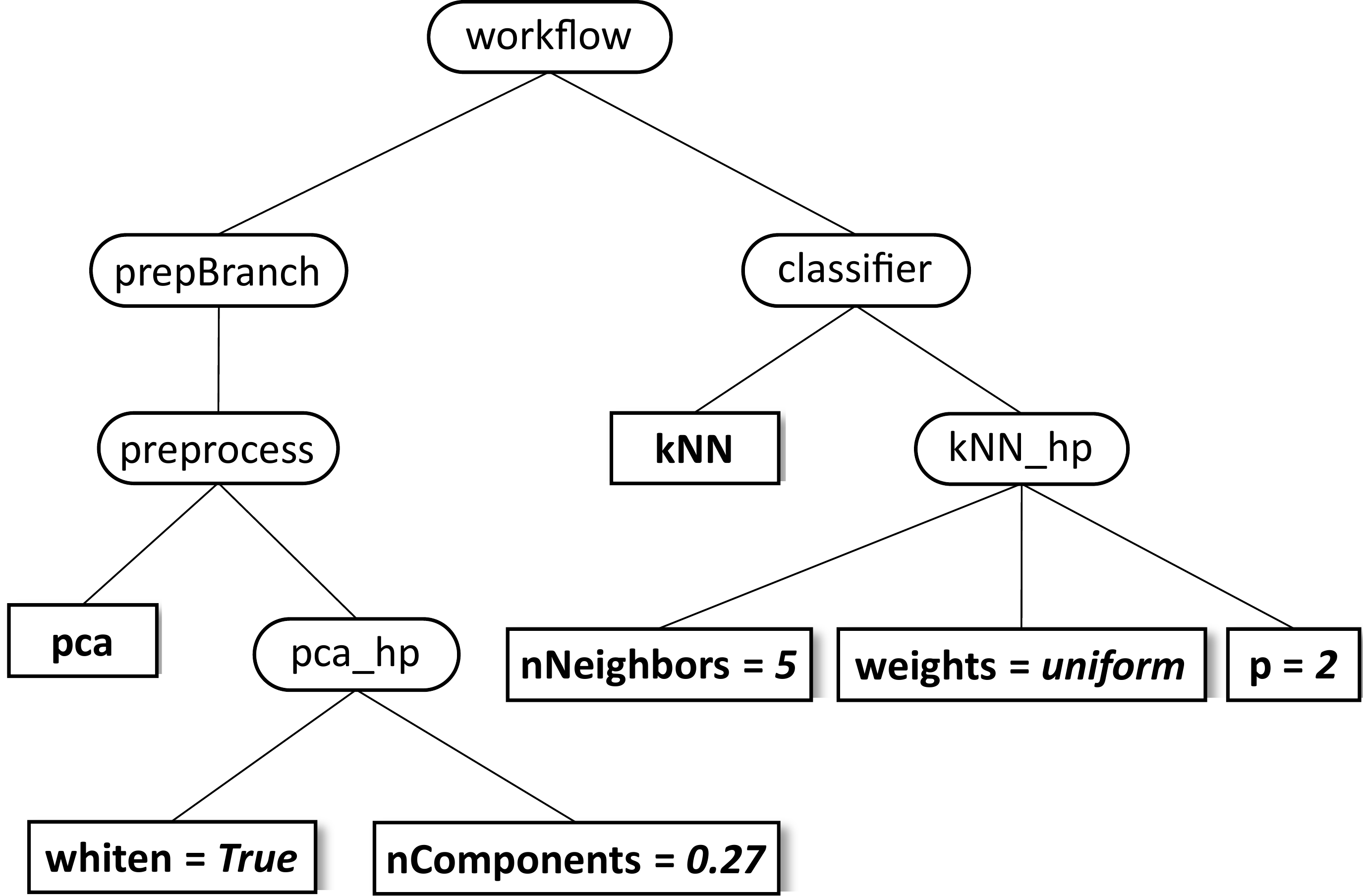}
    \caption{Genotype}
    \label{subfig:genotype}
\end{subfigure}

\begin{subfigure}{\linewidth}
    \centering
    \includegraphics[width=0.7\linewidth]{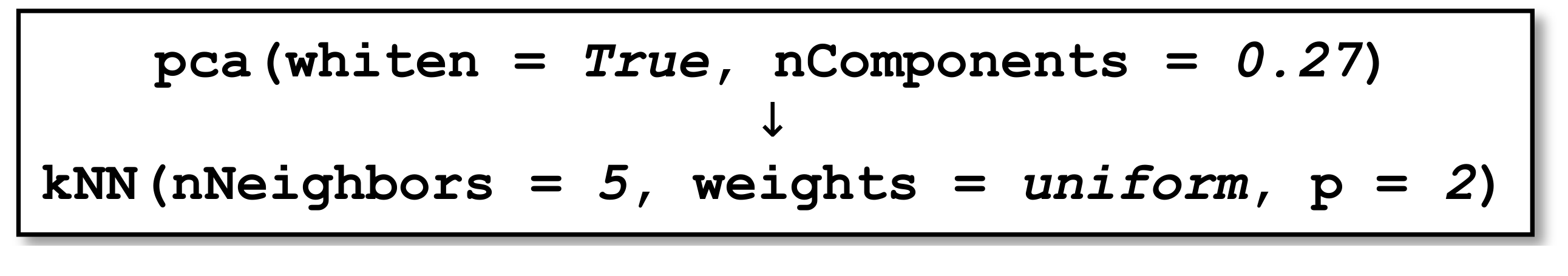}
    \caption{Phenotype}
    \label{subfig:phenotype}
\end{subfigure}
\caption{Example of individual}
\label{fig:encoding}
\end{figure}

The genotype of a valid individual is represented by its derivation tree of production rules, as formally defined by the \textit{cfg}. The phenotype represents the corresponding classification workflow. Figure~\ref{fig:encoding} illustrates an example mapping between genotype and phenotype. In this case, the dataset is preprocessed using principal component analysis to extract new features, followed by the k-nearest neighbour algorithm to build the classification model.

As mentioned in Section~\ref{subsec:backgound-ec}, the CFG defines the sets of terminal ($\sum_T$) and non-terminal ($\sum_N$) symbols, as well as the production rules (\textit{P}) for deriving valid expressions from the root symbol (\textit{S}). In the context of AWC, terminal symbols represent specific preprocessing and classification algorithms, along with their respective hyper-parameters. Non-terminal symbols define the derivable elements required to construct a classification workflow. Production rules dictate the derivation steps that generate valid workflows, which are ultimately expressed in terms of terminal symbols, as shown in Figure~\ref{subfig:phenotype}.

\begin{figure}[H]
\begin{center}
\begin{lstlisting}[xleftmargin=0.0\textwidth, xrightmargin=0.0\textwidth]
(*@$S$@*) = <workflow>

(*@$\sum_N$@*) = {(*@<workflow>,@*) (*@<prepBranch>,@*) (*@<classifier>,@*) (*@<preprocess>,@*) (*@<selectPercentile{\textunderscore}hp>,@*) (*@<rbfSampler{\textunderscore}hp>,@*) (*@<pca{\textunderscore}hp>,@*) (*@<normalizer{\textunderscore}hp>,@*) (*@<decisionTree{\textunderscore}hp>,@*) (*@<kNN{\textunderscore}hp>,@*) (*@<randomForest{\textunderscore}hp>,@*) (*@<gaussianNB{\textunderscore}hp>,@*) (*@<bernouilliNB{\textunderscore}hp>,@*)...}

(*@$\sum_T$@*) = {(*@\textit{selectPercentile}@*), (*@\textit{rbfSampler}@*), (*@\textit{pca}@*), (*@\textit{minMaxScaler}@*), (*@\textit{varianceThreshold}@*) (*@\textit{normalizer}@*) (*@\textit{percentile}@*), (*@\textit{gamma}@*), (*@\textit{nComponents}@*), (*@\textit{whiten}@*), (*@\textit{norm}@*), (*@\textit{decisionTree}@*),(*@\textit{kNN}@*), (*@\textit{randomForest}@*), (*@\textit{gaussianNB}@*), (*@\textit{bernouilliNB}@*), (*@\textit{criterion}@*), (*@\textit{maxDepth}@*), (*@\textit{maxFeatures}@*), (*@\textit{nNeighbors}@*), (*@\textit{weights}@*), (*@\textit{p}@*), (*@\textit{nEstimators}@*), (*@\textit{varSmoothing}@*), (*@\textit{alpha}@*), (*@\textit{fitPrior}@*), ...}

(*@$P$@*) = {
 <workflow>    ::= <prepBranch> <classifier> | <classifier>
 <prepBranch>  ::= <preprocess> | <prepBranch> <preprocess>

 <preprocess>  ::= (*@\textit{selectPercentile}@*) <selectPercentile_hp>
                 | (*@\textit{rbfSampler}@*) <rbfSampler_hp>
                 | (*@\textit{pca}@*) <pca_hp>
                 | (*@\textit{minMaxScaler}@*)
                 | (*@\textit{varianceThreshold}@*)
                 | (*@\textit{normalizer}@*) <normalizer_hp>
                 | ...

 <classifier>   ::= (*@\textit{decisionTree}@*) <decisionTree_hp>
                  | (*@\textit{kNN}@*) <kNN_hp>
                  | (*@\textit{randomForest}@*) <randomForest_hp>
                  | (*@\textit{gaussianNB}@*) <gaussianNB_hp>
                  | (*@\textit{bernouilliNB}@*) <bernouilliNB_hp>
                  | ...
                            
 <selectPercentile_hp>   ::= (*@\textit{percentile}@*)
 <rbfSampler_hp>         ::= (*@\textit{gamma}@*) (*@\textit{nComponents}@*)
 <pca_hp>                ::= (*@\textit{whiten}@*) (*@\textit{nComponents}@*)
 <normalizer_hp>         ::= (*@\textit{norm}@*)
 ...
                            
 <decisionTree_hp> ::= (*@\textit{criterion}@*) (*@\textit{maxDepth}@*) (*@\textit{maxFeatures}@*)... 
 <kNN_hp>          ::= (*@\textit{nNeighbors}@*) (*@\textit{weights}@*) (*@\textit{p}@*)
 <randomForest_hp> ::= (*@\textit{nEstimators}@*) (*@\textit{criterion}@*) (*@\textit{maxFeatures}@*)...
 <gaussianNB_hp>   ::= (*@\textit{varSmoothing}@*) 
 <bernouilliNB_hp> ::= (*@\textit{alpha}@*) (*@\textit{fitPrior}@*)
 ...
}

\end{lstlisting}
\end{center}
\caption{Grammar proposed for \ourtechnique}
\label{fig:grammar}
\end{figure}

Figure~\ref{fig:grammar} shows the proposed CFG for automated composition of classification workflows. For readability and space reasons, some symbols and production rules have been omitted (see supplementary material). The first two production rules in $P$ determine how the root symbol (\textit{$<$workflow$>$}) can be derived into a classifier (\textit{$<$classifier$>$}). In the first case, it can be preceded by a set of preprocessing methods represented by \textit{$<$prepBranch$>$}. The preprocessing branch allows for an inclusive sequence of preprocessing algorithms (\textit{$<$preprocess$>$}), covering various tasks such as feature extraction or feature selection. Notably, there are no restrictions on the type and sequence of preprocessing algorithms. On the other hand, \textit{$<$classifier$>$} can be derived into a classification algorithm and its corresponding hyper-parameters. It should be noted that adapting the grammar to other machine learning tasks, such as regression, can be achieved by directly adding the corresponding algorithms and hyper-parameters. Additionally, if additional constraints were imposed on workflows, the grammar could be adapted to consider specific algorithms, such as interpretable decision trees.

\subsection{Genetic operators}
\label{subsec:evoflow_operators}

Three genetic operators are employed in the evolutionary schema: selection, crossover, and mutation. The selection operator is applied at the beginning of each generation to choose a set of parents for breeding. It uses binary tournament selection, randomly selecting two individuals from \textit{pop} and taking the one with the best fitness. This process is repeated until \textit{popSize} individuals are selected.

\begin{figure}
\centering

\begin{subfigure}{\linewidth}
    \centering
    \includegraphics[width=\linewidth]{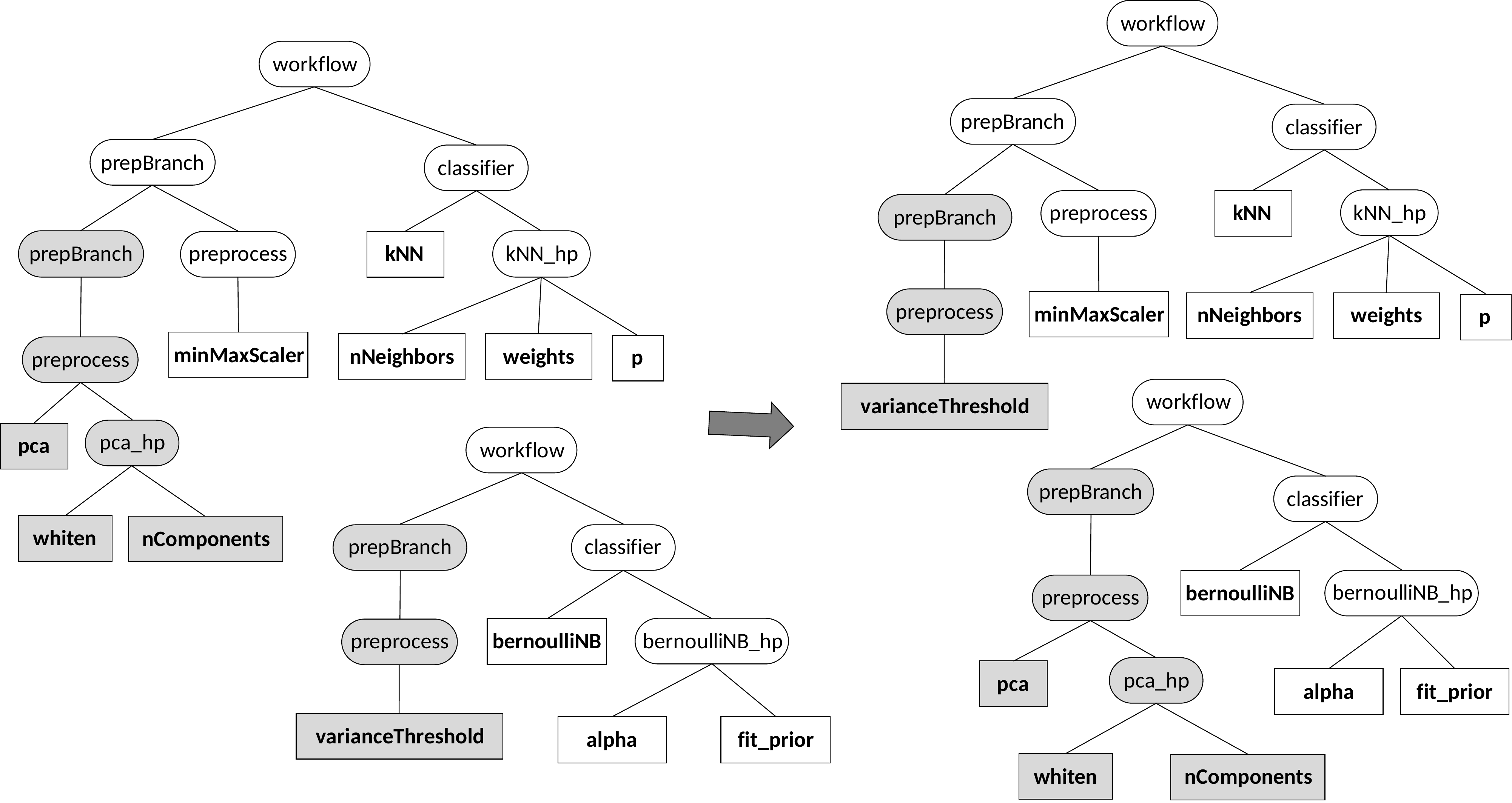}
    \caption{cxStruct}
    \label{subfig:cxStruct}
\end{subfigure}

\begin{subfigure}{\linewidth}
    \centering
    \includegraphics[width=\linewidth]{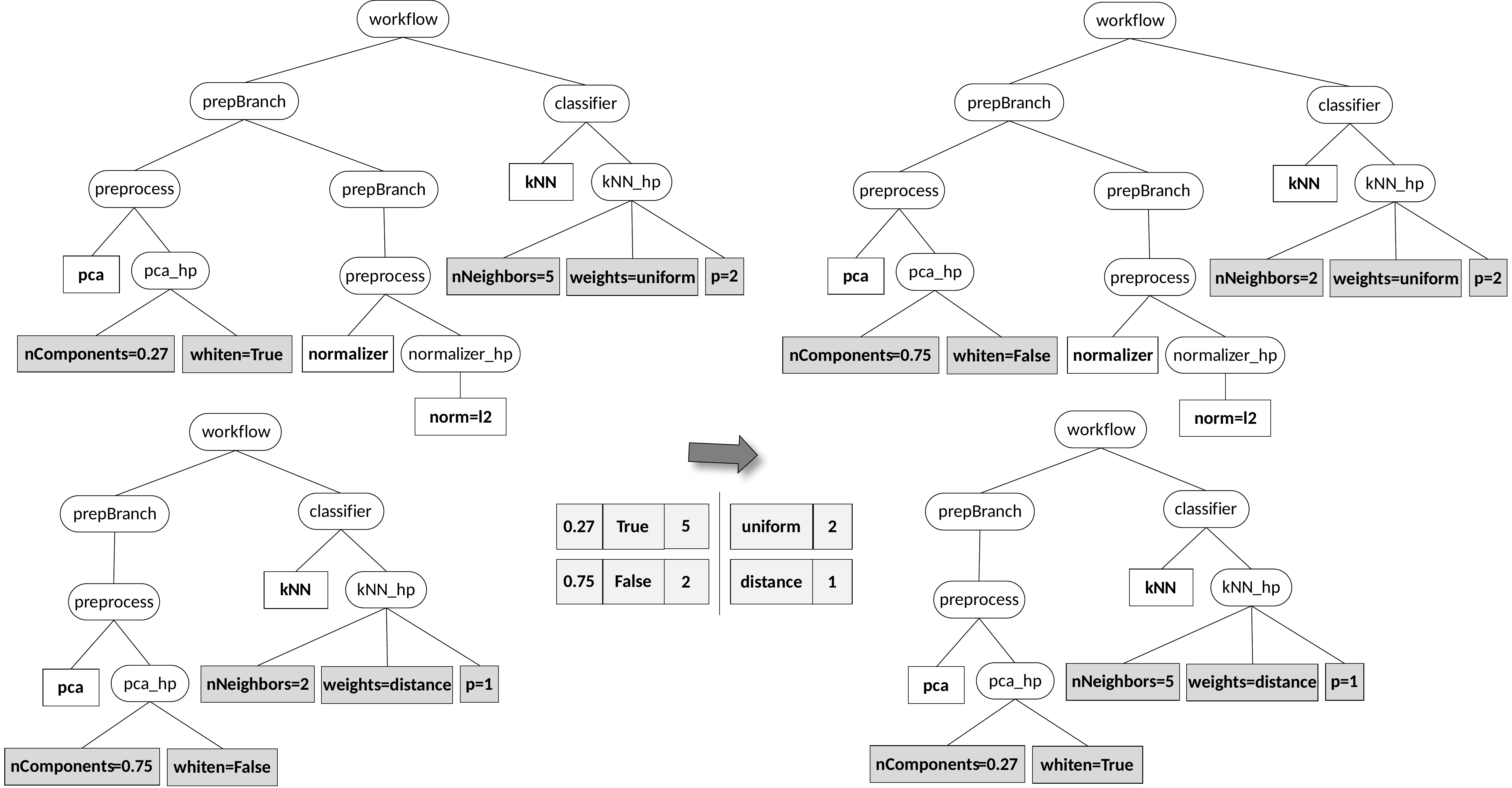}
    \caption{cxHparams}
    \label{subfig:cxHparams}
\end{subfigure}
\caption{Crossover example}
\label{fig:crossover}
\end{figure}

Crossover is applied to each pair of parents with a given probability. Two crossover operators are defined. First, \textit{cxStruct} is a classical GP operator that randomly selects a common non-terminal symbol in both parents to swap their respective subtrees. An example of this crossover is depicted in Figure~\ref{subfig:cxStruct}, with \textit{$<$prepBranch$>$} as the selected non-terminal symbol. Second, \textit{cxHparams} collects the common hyper-parameters of both parents and use them to compose two lists of the same length. A one-point crossover is then applied to swap the hyper-parameter values at a randomly selected crossover point. It is important to note that \textit{cxHparams} requires parents to share at least two hyper-parameters, making it inapplicable in some cases. Figure~\ref{subfig:cxHparams} depicts an example where only the hyper-parameters of the principal component analysis and k-nearest neighbours are eligible for swapping. Consequently, \ourtechnique gives preference to \textit{cxHparams} over \textit{cxStruct} when it is applicable.

\begin{figure}[H]
    \centering
     \begin{subfigure}{\textwidth}
        \centering
		 \includegraphics[scale=0.222]{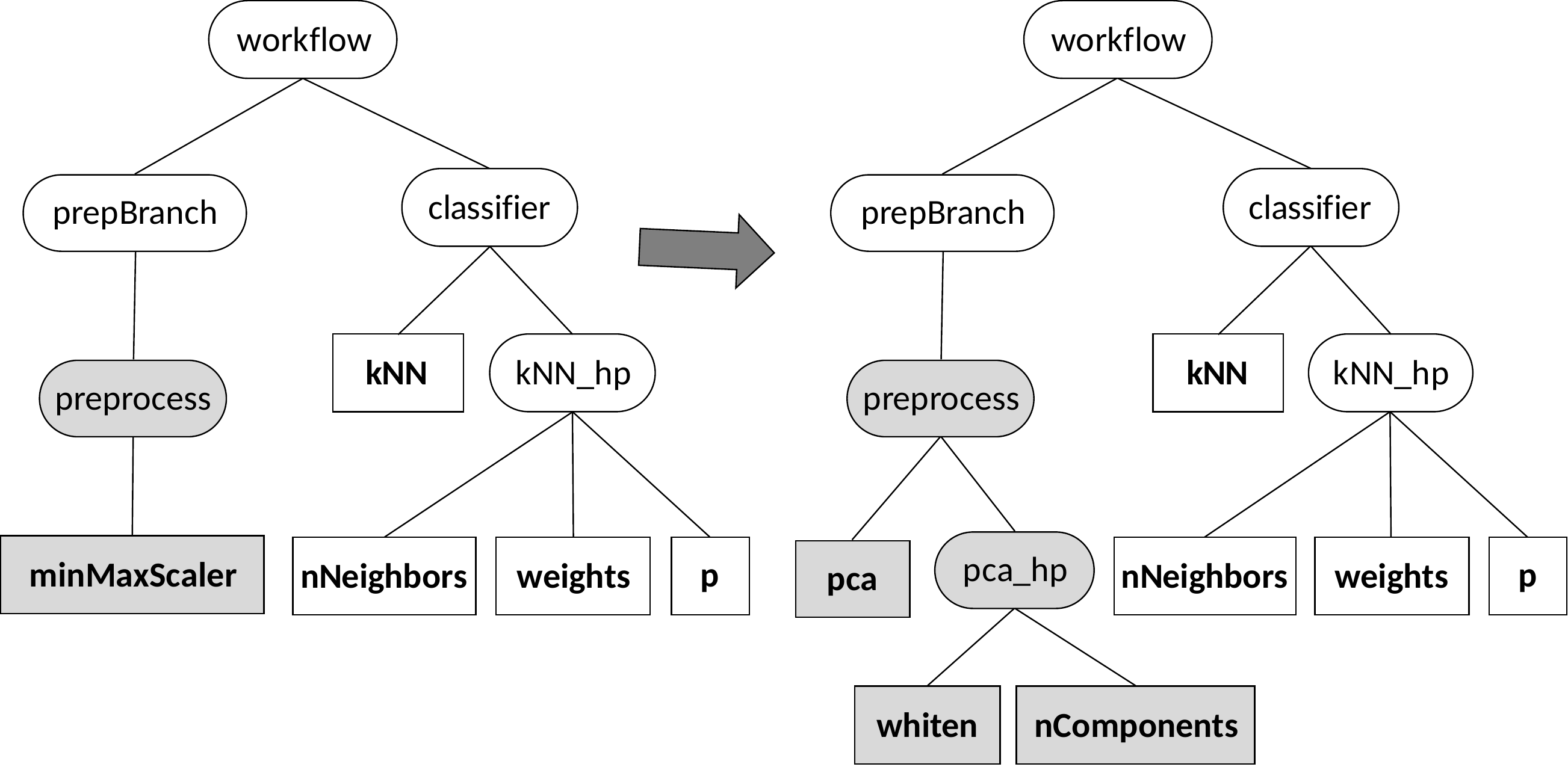}
        \caption{mutStruct}
         \label{subfig:mutStruct}
    \end{subfigure}
     \hfill
    \begin{subfigure}{\textwidth}
        \centering
        \includegraphics[scale=0.222]{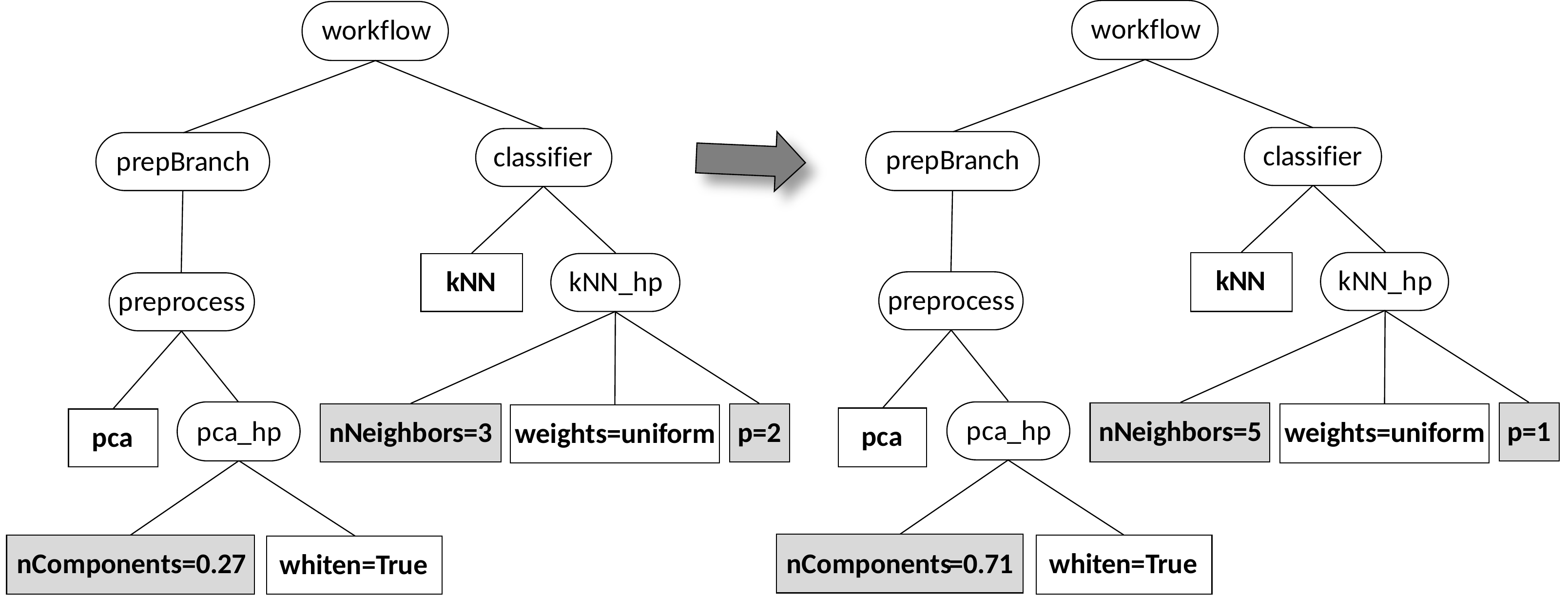}
        \caption{mutHparams}
        \label{subfig:mutHparams}
    \end{subfigure}
    \caption{Mutator example}
    \label{fig:mutation}
\end{figure}

Two variants of the mutation operator are considered. Firstly, \textit{mutStruct} aims to generate workflows with diverse algorithms. To this end, it randomly selects a non-terminal symbol and rebuilds the tree branch by deriving all non-terminal symbols randomly until only terminal symbols are generated. Figure~\ref{subfig:mutStruct} illustrates an example where the \textit{minMaxScaler} is removed and the \textit{pca} algorithm is added, along with its respective hyper-parameters. Secondly, \textit{mutHparams} randomly modifies the value of an hyper-parameter with a given probability, depending on whether it is related to a preprocessing or a classification algorithm. The probability of altering each preprocessing and classification hyper-parameter is calculated as the inverse of the number of preprocessing and classification hyper-parameters, respectively. For instance, the probability of modifying each preprocessing hyper-parameter in Figure~\ref{subfig:mutHparams} is 0.5, since there is only one preprocessing algorithm with two hyper-parameters, namely \textit{nComponents} and \textit{whiten}.

\subsection{Update procedure}
\label{subsec:evoflow_update}

After applying the genetic operators, the \textit{archive} is updated with \textit{archSize} individuals based on both their accuracy and workflow prediction diversity. When the \textit{archive} is empty in the first generation, the best \textit{archSize} individuals are directly added only based on their fitness. The workflow prediction diversity of each individual in the \textit{archive} is determined by comparing the predictions made during its evaluation with others. As mentioned above, cross-validation is used for evaluation, so the labels of all the samples in the \textit{train} set should be predicted for all the evaluated individuals. Consequently, these labels can be concatenated to build a prediction vector (\textit{x}) for each individual in \textit{archive}. 

Equation~\ref{ec:diversity} shows the computation of workflow prediction diversity (\textit{$div_i$}) for an individual \textit{i} in the \textit{archive}. Let \textit{$x_i$} be the prediction vector of individual \textit{i} defined as ($x_{i1}$, $x_{i2}$, ..., $x_{in}$), where \textit{n} corresponds to the number of samples in \textit{train}. Similarly, let \textit{$x_j$} be the prediction vector of another individual \textit{j} in \textit{archive} denoted as ($x_{j1}$, $x_{j2}$, ..., $x_{jn}$), where $i \neq j$. The workflow prediction diversity between these individuals is calculated by comparing \textit{$x_i$} and \textit{$x_j$} element-wise, counting the number of samples for which they make different predictions, as shown in Equation~\ref{ec:difference}. This count is then divided by the number of samples in \textit{train} to obtain the ratio of differences. This process is repeated for the remaining individuals in the \textit{archive}. The resulting ratios are aggregated and divided by the size of the \textit{archive} minus one. Thus, \textit{$div_i$} represents how dissimilar the predictions of individual \textit{i} are compared to the other individuals in the \textit{archive}, on average. Finally, a combined measure of workflow prediction diversity and fitness, denoted as \textit{divfit}, is calculated using the obtained value as shown in Equation~\ref{ec:combined}, where \textit{divWeight} is a parameter that determines the weight assigned to each component. This measure is used to sort the individuals in the \textit{archive}.

\begin{equation}
div_i = \frac{1}{|archive|-1} \mathop{\sum_{j=1}^{|archive|}}_{j \neq i} \frac{1}{|train|}\sum_{k=1}^{|train|} diff (x_{
ik}, x_{jk})
\label{ec:diversity}
\end{equation}

\begin{equation}
diff(x,y) =
\begin{cases} 
      1 & \text{if $x = y$}, \\
      0 & \text{otherwise}
   \end{cases}
\label{ec:difference}
\end{equation}

\begin{equation}
divfit_i = divWeight*div_i+(1-divWeight)*fitness_i
\label{ec:combined}
\end{equation}

In subsequent generations, when the \textit{archive} is not empty as it was after evaluating the initial population, the update procedure slightly differs. It begins by selecting those individuals in \textit{pop} with a fitness greater than 0. Then, for each individual, it computes how diverse its predictions are compared to those individuals already in the \textit{archive}. Since this individual is not in the \textit{archive}, two modifications are made to Equation~\ref{ec:diversity} to compute its \textit{div} value: (1) the denominator of the outer division is limited to $|archive|$; and (2) the condition $j \neq i$ is eliminated as \textit{i} and \textit{j} will never refer to the same individual. After computing \textit{divfit}, individuals are added to \textit{archive}, maintaining the sorted order. Finally, the last individuals are trimmed from \textit{archive} to retain only the top \textit{archSize} individuals with the highest \textit{divfit} value.

\section{Experimental settings}
\label{sec:experiments}

An implementation of \ourtechnique is provided in Python using the DEAP framework (\textit{Distributed Evolutionary Algorithms in Python})~\citep{fortin2012}, which offers functionalities and data structures for implementing evolutionary algorithms. Preprocessing algorithms were obtained from imbalanced-learn\footnote{\textit{imbalanced-learn}, available at \url{https://imbalanced-learn.org/} (Last accessed: January 19, 2024)} and scikit-learn. For classification algorithms, scikit-learn and XGBoost\footnote{\textit{xgboost}, available at \url{https://xgboost.ai/} (Last accessed: January 19, 2024)} were considered. Implementations are available for download from the supplementary material. 

In this section, we first outline the research questions (RQs) that drive our objectives. Then, we describe the methodology employed in the experiments.

\subsection{Research Questions}
\label{subsec:rquestions}

The conducted experiments aim to address the following RQs:

\begin{itemize}
    \item \textit{RQ1. How do the AWC-specific genetic operators and ensembles contribute to EvoFlow’s model?} Given that \ourtechnique incorporates two mechanisms designed specifically for the AWC problem, it is essential to assess the individual and combined impact of these mechanisms on the predictive performance of the final models.
    
    \item \textit{RQ2. How does EvoFlow's effectiveness compare to other AWC approaches using different techniques?} As mentioned in Section~\ref{sec:related}, various proposals have effectively tackled the AWC problem using different techniques. it is crucial to conduct a comprehensive comparison between \ourtechnique, based on G3P, and these existing approaches to determine if it achieves state-of-the-art performance.

    \item \textit{RQ3. How does EvoFlow compare in effectiveness to another GP3-based AWC proposal?}  Similarly to RQ2, it is valuable to analyse the benefits that \ourtechnique brings in comparison to other G3P-based AWC tools, ensuring that it advances the use of grammar-based methods for the AWC problem.
\end{itemize}

\subsection{Experimental framework}
\label{subsec:experiments_setup}

\begin{table}[H]
\small\footnotesize
\centering
\begin{tabular}{lcccc}
\hline
 & Features & Classes & Train & Test \\
\hline
abalone & 8 &26 &2,924 &1,253 \\
amazon &10,000 &50 &1,050 &450 \\
breastcancer* & 9 & 2 & 466 & 233 \\
car &6 &4 &1,210 &518 \\
convex &784 &2 &8,000 &50,000 \\
dexter &20,000 &2 &420 &180 \\
dorothea &100,000 &2 &805 &345 \\
germancredit &20 &2 &700 &300 \\
gisette &5,000 &2 &4,900 &2,100 \\
glass* & 9 & 6 & 142 & 72 \\
hillvalley* & 100 & 2 & 808 & 404 \\
ionosphere* & 33 & 2 & 234 & 117 \\
madelon &500 &2 &1,820 &780 \\
krvskp &36 &2 &2,238 &958 \\
secom &590 &2 &1,097 &470 \\
semeion &256 &10 &1,116 &477 \\
shuttle &9 &7 &43,500 &14,500 \\
spambase* & 57 & 2 & 3,067 & 1,534 \\
waveform &40 &3 &3,500 &1,500 \\
winered* & 11 & 6 & 1,066 & 533 \\
winewhite &11 &7 &3,429 &1,469 \\
yeast &8 &10 &1,039 &445 \\
\hline
\end{tabular}
\caption{Datasets used for experiments}
\label{tab:datasets}
\end{table}

For the experimentation, we selected a total of twenty-two classification datasets from the literature~\citep{thornton2013, de2017}. Table~\ref{tab:datasets} provides details on each dataset, including the number of features and classes, as well as the sizes of the training and test sets. These datasets offer diversity in terms of their sizes and complexities. For datasets sources from~\citep{thornton2013}, we used the same data partition defined by the authors. Datasets taken from~\citep{de2017}, marked with an asterisk (*), were randomly partitioned, with one-third of the samples allocated for validation, to maintain consistent conditions across all datasets. The generated partitions are publicly available as supplementary material, ensuring reproducibility.

Table~\ref{tab:exeCfg} shows the parameter values for \ourtechnique. The population size, number of generations, and crossover and structural mutation probabilities were determined through preliminary experiments. For each execution, 1, 6 and 12 hours of budget were assigned to \ourtechnique and the baseline methods, with the 12-hour budget used only for datasets where significant differences were observed with the other budgets. The evaluation budget was set to one-tenth of the total budget to prevent spending the execution time on a single individual. The maximum number of derivations, which governs the genotype size, was set to 13. This setting enables the creation of workflows comprising a sequence of up to five algorithms, in accordance with the proposed grammar. The archive size, representing the ensemble size, was limited to 10 to minimise the overhead of retraining the workflows with the full training set after \ourtechnique completes. Finally, the diversity weight of 0.2 was chosen based on preliminary experiments, as higher values tend to yield ensembles with workflows exhibiting lower predictive performance.

\begin{table}[H]
    \small\footnotesize
	\centering
	\setlength{\tabcolsep}{15pt}
	\begin{tabular}{c  c} 
	    \hline
		Parameter & Value \\ 
		\hline
		Number of generations \textit{(maxGen)} & 100 \\
        Population size \textit{(popSize)} & 100 \\
        Crossover probability (\textit{cxProb)} & 0.8 \\
        Structural mutation probability \textit{(stMutProb)} & 0.2 \\
        Budget (hours) & [1, 6, 12] \\
        Evaluation budget (minutes) & [6, 36, 72] \\
        Maximum number of derivations \textit{(maxDer)} & 13 \\
        Size of the archive \textit{(archSize)} & 10 \\
        Diversity weight (\textit{divWeight}) & 0.2 \\
        \hline
	\end{tabular}
	\caption{Parameter setting of \ourtechnique}
	\label{tab:exeCfg}
\end{table}

To address the RQs, three experiments were conducted. The first experiment (Section~\ref{subsec:experiments_internals}) analyses the impact of specific genetic operators and diverse ensembles on the predictive performance of \ourtechnique. Our results are then compared with those of Auto-Sklearn~\citep{feurer2020}, TPOT~\citep{olson2016}, and ML-Plan~\citep{mohr2018} in the second experiment (Section~\ref{subsec:experiments_comparison}). These approaches employ BO, GP and AI planning, respectively. Finally, a third experiment (Section~\ref{sec:experiments_recipe}) compares \ourtechnique with RECIPE~\citep{de2017}, another G3P-based proposal for AWC. This separate comparison is necessary because RECIPE uses an older version of \textit{scikit-learn} that does not compute the balanced accuracy score, which is the default measure employed by the publicly available implementations of TPOT and Auto-Sklearn at the time of this experimentation. Thus, the $F_1$ score, representing the harmonic mean of precision and recall, was used as the fitness measure for RECIPE. For \ourtechnique, the solutions obtained from the previous experiment, optimised based on the balanced accuracy score, were used, although the $F_1$ score was computed for the test sets in this experiment. The hyper-parameters of the baseline methods were set to their default values, with only the budget being modified as indicated above. To ensure result validity, twenty repetitions were conducted using different random seeds. The raw results, along with the experiment scripts, are publicly available as supplementary material.

\section{Experiment 1: Ablation study}
\label{subsec:experiments_internals}

To address RQ1, this section analyses the internal mechanisms of \ourtechnique, specifically examining the impact of using specific genetic operators and constructing ensembles consisting of diverse workflows. Four versions of the method are considered: (1) basic-\ourtechnique, which employs standard GP operators and returns the best workflow; (2) op-\ourtechnique, which uses only specific genetic operators; (3) ens-\ourtechnique, which focuses solely on building diverse ensembles; and (4) \ourtechnique, the complete proposed method that incorporates both specific genetic operators and diverse ensembles. The experiment is conducted with a budget of 1 hour.

\begin{table}[H]
\small\footnotesize
\centering
\begin{tabular}{lccc}
\hline
                 & \textbf{basic-\ourtechnique} & \textbf{op-\ourtechnique} & \textbf{Inc/Dec (\%)} \\
\hline
abalone          & $0.1467 \pm  0.0087$         & $0.1487 \pm 0.0149$     & 1.3880                 \\
amazon           & $0.8425 \pm  0.0083$         & $0.8404 \pm 0.0084$     & -0.2422                \\
breastcancer     & $0.9824 \pm  0.0068$         & $0.9823 \pm 0.0056$     & -0.0108                \\
car              & $0.9965 \pm  0.0118$         & $0.9962 \pm 0.0098$     & -0.0286                \\
convex           & $0.7715 \pm  0.0093$         & $0.7703 \pm 0.0100$     & -0.1562                \\
dexter           & $0.9481 \pm  0.0124$         & $0.9453 \pm 0.0141$     & -0.3037                \\
dorothea         & $0.7950 \pm  0.0508$         & $0.7934 \pm 0.0493$     & -0.2024                \\
germancredit     & $0.7036 \pm  0.0171$         & $0.6894 \pm 0.0186$     & -2.0158                \\
gisette          & $0.9754 \pm  0.0024$         & $0.9754 \pm 0.0024$     & 0.0000                 \\
glass            & $0.7484 \pm  0.0712$         & $0.7482 \pm 0.0262$     & -0.0196                \\
hillvalley       & $0.9995 \pm  0.0010$         & $0.9996 \pm 0.0009$     & 0.0124                 \\
ionosphere       & $0.9517 \pm  0.0147$         & $0.9529 \pm 0.0196$     & 0.1226                 \\
krvskp           & $0.9924 \pm  0.0048$         & $0.9940 \pm 0.0017$     & 0.1583                 \\
madelon          & $0.8543 \pm  0.0330$         & $0.8595 \pm 0.0184$     & 0.6007                 \\
secom            & $0.6143 \pm  0.0413$         & $0.6061 \pm 0.0403$     & -1.3324                \\
semeion          & $0.9216 \pm  0.0206$         & $0.9230 \pm 0.0157$     & 0.1495                 \\
shuttle          & $0.9907 \pm  0.0042$         & $0.9904 \pm 0.0036$     & -0.0316                \\
spambase         & $0.9530 \pm  0.0029$         & $0.9545 \pm 0.0029$     & 0.1574                 \\
waveform         & $0.8602 \pm  0.0040$         & $0.8568 \pm 0.0056$     & -0.3920                \\
winered          & $0.3554 \pm  0.0223$         & $0.3649 \pm 0.0283$     & 2.6712                 \\
winewhite        & $0.3950 \pm  0.0571$         & $0.4036 \pm 0.0604$     & 2.1782                 \\
yeast            & $0.5615 \pm  0.0291$         & $0.5732 \pm 0.0238$     & 2.0985 \\
\hline
\end{tabular}
\caption{basic-\ourtechnique compared to op-\ourtechnique in terms of their balanced accuracy score}
\label{tab:op-evoflow}
\end{table}

First, we compare basic-\ourtechnique with op-\ourtechnique. Table~\ref{tab:op-evoflow} presents the average value obtained for each version. The ``Inc/dec (\%)'' column indicates the percentage increase (positive value) or decrease (negative value) achieved by op-\ourtechnique compared to basic-\ourtechnique. As observed, both versions yield similar results, with the largest difference occurring in the \textit{winered} dataset. We also perform a Wilcoxon signed-rank test on the average values across twenty repetitions for each dataset. At a significance level of 0.05, no significant differences in favour of any version are found.

\begin{table}[H]
\small\footnotesize
\centering
\begin{tabular}{lccccc}
\hline
             & \textbf{basic-\ourtechnique} & \textbf{ens-\ourtechnique} & \textbf{top10-\ourtechnique} & \textbf{top10W-\ourtechnique} \\
\hline
abalone      & $0.1467 \pm  0.0087$                 & $\textbf{0.1468} \pm 0.0094$             & $0.1448 \pm 0.0092$                      & $0.1434 \pm 0.0087$                       \\
amazon       & $0.8425 \pm  0.0083$                 & $\textbf{0.8447} \pm 0.0105$             & $0.8400 \pm 0.0126$                      & $0.8422 \pm 0.0097$                       \\
breastcancer & $0.9824 \pm  0.0068$                 & $\textbf{0.9841} \pm 0.0065$             & $0.9837 \pm 0.0072$                      & $0.9837 \pm 0.0072$                       \\
car          & $0.9965 \pm  0.0118$                 & $\textbf{0.9972} \pm 0.0090$             & $0.9964 \pm 0.0094$                      & $0.9964 \pm 0.0094$                       \\
convex       & $0.7715 \pm  0.0093$                 & $\textbf{0.7783} \pm 0.0133$             & $0.7720 \pm 0.0183$                      & $0.7756 \pm 0.0154$                       \\
dexter       & $0.9481 \pm  0.0124$                 & $0.9540 \pm 0.0075$                      & $\textbf{0.9554} \pm 0.0069$             & $0.9549 \pm 0.0077$                       \\
dorothea     & $\textbf{0.7950} \pm  0.0508$        & $0.7666 \pm 0.0492$                      & $0.7395 \pm 0.0406$                      & $0.7666 \pm 0.0492$                       \\
germancredit & $\textbf{0.7036} \pm  0.0171$        & $0.7018 \pm 0.0147$                      & $0.7027 \pm 0.0174$                      & $0.7015 \pm 0.0178$                       \\
gisette      & $0.9754 \pm  0.0024$                 & $\textbf{0.9767} \pm 0.0020$             & $\textbf{0.9767} \pm 0.0018$             & $0.9766 \pm 0.0020$                       \\
glass        & $0.7484 \pm  0.0712$                 & $\textbf{0.7541} \pm 0.0549$             & $0.7476 \pm 0.0655$                      & $0.7473 \pm 0.0655$                       \\
hillvalley   & $0.9995 \pm  0.0010$                 & $\textbf{1.0000} \pm 0.0000$             & $\textbf{1.0000} \pm 0.0000$             & $\textbf{1.0000} \pm 0.0000$              \\
ionosphere   & $0.9517 \pm  0.0147$                 & $\textbf{0.9565} \pm 0.0175$             & $0.9516 \pm 0.0171$                      & $0.9519 \pm 0.0170$                        \\
krvskp       & $0.9924 \pm  0.0048$                 & $\textbf{0.9927} \pm 0.0043$             & $0.9923 \pm 0.0043$                      & $0.9924 \pm 0.0043$                        \\
madelon      & $\textbf{0.8543} \pm  0.0330$        & $0.8542 \pm 0.0275$                      & $0.8445 \pm 0.0376$                      & $0.8489 \pm 0.0378$                        \\
secom        & $0.6143 \pm  0.0413$                 & $\textbf{0.6195} \pm 0.0472$             & $0.6143 \pm 0.0492$                      & $0.6150 \pm 0.0504$                        \\
semeion      & $\textbf{0.9216} \pm  0.0206$        & $0.9209 \pm 0.0147$                      & $0.9171 \pm 0.0162$                      & $0.9173 \pm 0.0166$                        \\
shuttle      & $\textbf{0.9907} \pm  0.0042$        & $0.9850 \pm 0.0137$                      & $0.9839 \pm 0.0135$                      & $0.9884 \pm 0.0045$                        \\
spambase     & $\textbf{0.9530} \pm  0.0029$        & $\textbf{0.9530} \pm 0.0035$             & $0.9522 \pm 0.0030$                      & $0.9525 \pm 0.0031$                        \\
waveform     & $0.8602 \pm  0.0040$                 & $\textbf{0.8613} \pm 0.0032$             & $0.8588 \pm 0.0037$                      & $0.8596 \pm 0.0041$                        \\
winered      & $0.3554 \pm  0.0223$                 & $\textbf{0.3602} \pm 0.0173$             & $0.3533 \pm 0.0236$                      & $0.3528 \pm 0.0223$                        \\
winewhite    & $0.3950 \pm  0.0571$                 & $0.4032 \pm 0.0441$                      & $0.4076 \pm 0.0527$                      & $\textbf{0.4137} \pm 0.0584$               \\
yeast        & $0.5615 \pm  0.0291$                 & $\textbf{0.5711} \pm 0.0190$             & $0.5634 \pm 0.0303$                      & $0.5635 \pm 0.0302$  \\     
\hline
\end{tabular}
\caption{Analysis of diverse ensembles in terms of their balanced accuracy score}
\label{tab:ens-evoflow}
\end{table}

As mentioned in Section~\ref{sec:related}, the practice of combining the best workflows into an ensemble has already been employed in the AWC literature. Thus, comparing basic-\ourtechnique with ens-\ourtechnique alone might obscure the actual contribution of building diverse ensembles, as the improvement could merely come from having an ensemble. To address this concern, we examine two additional versions: top10-\ourtechnique and top10W-\ourtechnique. These versions create an ensemble comprising the top ten workflows based solely on their predictive performance. However, top10W-\ourtechnique constructs a weighted ensemble, assigning greater importance to workflows with better fitness. Notice that diversity is not considered for these versions. Table~\ref{tab:ens-evoflow} presents their balanced accuracy scores. As observed, ens-\ourtechnique achieves the highest balanced accuracy score in thirteen out of twenty-two datasets, while basic-\ourtechnique outperforms it in five datasets. Regarding top10-\ourtechnique and top10W-\ourtechnique, they only yield the best results in two and one dataset(s), respectively. Additionally, we perform the Friedman test to check for significant differences in average values across twenty-two datasets. After rejecting the null hypothesis, the post hoc Holm test reveals that ens-\ourtechnique outperforms the other versions with a significance level of 0.05. It is worth noting that we chose the Wilcoxon signed-rank test and the Friedman test due to their robustness in non-parametric settings, essential given the complex nature of algorithmic performance in AutoML. As stated by García et al.~\cite{garcia2010}, the Wilcoxon test is effective in dealing with paired samples for comparing different versions of our technique, and the Friedman test is particularly suited for evaluating multiple algorithms across various datasets. This ensures reliable statistical validation, even in the presence of outliers and small sample sizes.

Having established the effectiveness of building diverse workflow ensembles, we proceed to compare \ourtechnique with the aforementioned versions. We conduct statistical tests on the results presented in Tables~\ref{tab:op-evoflow} and~\ref{tab:ens-evoflow}, showing that \ourtechnique significantly outperforms basic-\ourtechnique, op-\ourtechnique, and ens-\ourtechnique in four, five and one dataset(s), respectively. None of the other versions can outperform \ourtechnique in any dataset. Furthermore, we compare these versions based on their average values across datasets. In this regard, \ourtechnique consistently performs significantly better than the other options with a significance level of 0.05, thus demonstrating the beneficial combination of specific genetic operators and diverse workflow ensembles. Therefore, this version is henceforth used in the subsequent comparisons.

\section{Experiment 2: \ourtechnique compared to other AutoML proposals}
\label{subsec:experiments_comparison}

\begin{sidewaystable}[!hpt]
\footnotesize
\centering
\begin{tabular}{llllllllll}
\hline
& \multicolumn{4}{c}{\textbf{1-hour budget}} && \multicolumn{4}{c}{\textbf{6-hour budget}} \\
\hline
 & \textbf{\ourtechnique} & \textbf{Auto-Sklearn} & \textbf{TPOT} & \textbf{ML-Plan} && \textbf{\ourtechnique} & \textbf{Auto-Sklearn} & \textbf{TPOT} & \textbf{ML-Plan} \\
\cline{2-5} \cline{7-10}

abalone & $0.1539 \pm 0.0161$ & $0.1469 \pm 0.0076$ & $0.1389 \pm 0.0149$ & $0.1520 \pm 0.0090$ && $0.1560 \pm 0.0127$ & $0.1442 \pm 0.0128$ & $0.1401 \pm 0.0133\blacktriangledown$ & $0.1553 \pm 0.0037$ \\
amazon & $0.8425 \pm 0.0128$ & $0.8653 \pm 0.0110\blacktriangle$ & $0.7544 \pm 0.0511\blacktriangledown$ & $0.7126 \pm 0.0000\blacktriangledown$ && $0.8520 \pm 0.0087$ & $0.8137 \pm 0.0423\blacktriangledown$ & $0.7888 \pm 0.0323\blacktriangledown$ & $0.7153 \pm 0.0065\blacktriangledown$ \\
breastcancer & $0.9831 \pm 0.0058$ & $0.9801 \pm 0.0061$ & $0.9812 \pm 0.0053$ & $0.9865 \pm 0.0042$ && $0.9829 \pm 0.0066$ & $0.9802 \pm 0.0067$ & $0.9771 \pm 0.0089$ & $0.9872 \pm 0.0010$ \\
car & $0.9991 \pm 0.0037$ & $0.9464 \pm 0.0602\blacktriangledown$ & $0.9061 \pm 0.0744\blacktriangledown$ & $0.9703 \pm 0.0200\blacktriangledown$ && $0.9977 \pm 0.0055$ & $0.9908 \pm 0.0135$ & $0.9009 \pm 0.0813\blacktriangledown$ & $0.9836 \pm 0.0150\blacktriangledown$ \\
convex & $0.7808 \pm 0.0141$ & $0.7615 \pm 0.0012\blacktriangledown$ & $0.7727 \pm 0.0184$ & $0.7510 \pm 0.0065\blacktriangledown$ && $0.8035 \pm 0.0102$ & $0.7590 \pm 0.0044\blacktriangledown$ & $0.7935 \pm 0.0137\blacktriangledown$ & $0.7518 \pm 0.0015\blacktriangledown$ \\
dexter & $0.9541 \pm 0.0073$ & $0.9257 \pm 0.0103\blacktriangledown$ & $0.9360 \pm 0.0090\blacktriangledown$ & $0.9118 \pm 0.0169\blacktriangledown$ && $0.9447 \pm 0.0099$ & $0.9113 \pm 0.0503\blacktriangledown$ & $0.9343 \pm 0.0099\blacktriangledown$ & $0.9285 \pm 0.0213\blacktriangledown$ \\
dorothea & $0.7709 \pm 0.0460$ & $0.7242 \pm 0.0000\blacktriangledown$ & $0.7090 \pm 0.0430\blacktriangledown$ & $0.7415 \pm 0.0475$ && $0.8176 \pm 0.0259$ & $0.7332 \pm 0.0254\blacktriangledown$ & $0.7214 \pm 0.0556\blacktriangledown$ & $0.7471 \pm 0.0595$ \\
germancredit & $0.6920 \pm 0.0187$ & $0.6947 \pm 0.0139$ & $0.6729 \pm 0.0062\blacktriangledown$ & $0.6395 \pm 0.0250\blacktriangledown$ && $0.6936 \pm 0.0178$ & $0.6973 \pm 0.0165$ & $0.6741 \pm 0.0129\blacktriangledown$ & $0.6552 \pm 0.0263\blacktriangledown$ \\
gisette & $0.9767 \pm 0.0021$ & $0.9719 \pm 0.0025\blacktriangledown$ & $0.9722 \pm 0.0041\blacktriangledown$ & $0.9738 \pm 0.0067$ && $0.9793 \pm 0.0018$ & $0.9778 \pm 0.0009$ & $0.9771 \pm 0.0024\blacktriangledown$ & $0.9763 \pm 0.0032\blacktriangledown$ \\
glass & $0.7562 \pm 0.0183$ & $0.6965 \pm 0.0702\blacktriangledown$ & $0.7168 \pm 0.0410\blacktriangledown$ & $0.6976 \pm 0.0433\blacktriangledown$ && $0.7519 \pm 0.0310$ & $0.7260 \pm 0.0574$ & $0.7236 \pm 0.0472$ & $0.6919 \pm 0.0418\blacktriangledown$ \\
hillvalley & $1.0000 \pm 0.0000$ & $0.9588 \pm 0.0092\blacktriangledown$ & $0.9861 \pm 0.0134\blacktriangledown$ & $0.9620 \pm 0.0052\blacktriangledown$ && $1.0000 \pm 0.0000$ & $0.9542 \pm 0.0182\blacktriangledown$ & $0.9993 \pm 0.0020$ & $0.9603 \pm 0.0059\blacktriangledown$ \\
ionosphere & $0.9599 \pm 0.0149$ & $0.9330 \pm 0.0081\blacktriangledown$ & $0.9585 \pm 0.0116$ & $0.9501 \pm 0.0182$ && $0.9606 \pm 0.0167$ & $0.9363 \pm 0.0079\blacktriangledown$ & $0.9603 \pm 0.0103$ & $0.9533 \pm 0.0165$ \\
krvskp & $0.9945 \pm 0.0007$ & $0.9942 \pm 0.0019$ & $0.9923 \pm 0.0041$ & $0.9951 \pm 0.0030$ && $0.9947 \pm 0.0006$ & $0.9948 \pm 0.0015$ & $0.9935 \pm 0.0023$ & $0.9957 \pm 0.0019$ \\
madelon & $0.8568 \pm 0.0226$ & $0.8205 \pm 0.0089\blacktriangledown$ & $0.8486 \pm 0.0074$ & $0.7796 \pm 0.0168\blacktriangledown$ && $0.8895 \pm 0.0076$ & $0.8256 \pm 0.0075\blacktriangledown$ & $0.8504 \pm 0.0116\blacktriangledown$ & $0.7831 \pm 0.0211\blacktriangledown$ \\
secom & $0.6176 \pm 0.0327$ & $0.5349 \pm 0.0386\blacktriangledown$ & $0.5403 \pm 0.0402\blacktriangledown$ & $0.5743 \pm 0.0480$ && $0.6192 \pm 0.0236$ & $0.5702 \pm 0.0365\blacktriangledown$ & $0.5447 \pm 0.0507\blacktriangledown$ & $0.5750 \pm 0.0431\blacktriangledown$ \\
semeion & $0.9263 \pm 0.0096$ & $0.9006 \pm 0.0052\blacktriangledown$ & $0.9228 \pm 0.0097$ & $0.9251 \pm 0.0080$ && $0.9294 \pm 0.0130$ & $0.8755 \pm 0.0103\blacktriangledown$ & $0.9285 \pm 0.0067$ & $0.9274 \pm 0.0147$ \\
shuttle & $0.9912 \pm 0.0049$ & $0.9872 \pm 0.0081$ & $0.9392 \pm 0.0539\blacktriangledown$ & $0.9305 \pm 0.0345\blacktriangledown$ && $0.9912 \pm 0.0045$ & $0.9889 \pm 0.0036$ & $0.9665 \pm 0.0407\blacktriangledown$ & $0.9137 \pm 0.0637\blacktriangledown$ \\
spambase & $0.9552 \pm 0.0032$ & $0.9506 \pm 0.0127$ & $0.9547 \pm 0.0042$ & $0.9554 \pm 0.0004$ && $0.9564 \pm 0.0025$ & $0.9565 \pm 0.0047$ & $0.9562 \pm 0.0035$ & $0.9557 \pm 0.0045$ \\
waveform & $0.8608 \pm 0.0025$ & $0.8435 \pm 0.0069\blacktriangledown$ & $0.8556 \pm 0.0027\blacktriangledown$ & $0.8564 \pm 0.0012\blacktriangledown$ && $0.8606 \pm 0.0026$ & $0.8464 \pm 0.0083\blacktriangledown$ & $0.8581 \pm 0.0041$ & $0.8579 \pm 0.0025\blacktriangledown$ \\
winered & $0.3718 \pm 0.0249$ & $0.3386 \pm 0.0362\blacktriangledown$ & $0.3252 \pm 0.0249\blacktriangledown$ & $0.2905 \pm 0.0212\blacktriangledown$ && $0.3694 \pm 0.0196$ & $0.3747 \pm 0.0249$ & $0.3386 \pm 0.0249\blacktriangledown$ & $0.2945 \pm 0.0176\blacktriangledown$ \\
winewhite & $0.3927 \pm 0.0091$ & $0.3979 \pm 0.0930$ & $0.3308 \pm 0.0770\blacktriangledown$ & $0.3659 \pm 0.0080\blacktriangledown$ && $0.4080 \pm 0.0580$ & $0.3798 \pm 0.0576$ & $0.3473 \pm 0.0623\blacktriangledown$ & $0.3625 \pm 0.0217\blacktriangledown$ \\
yeast & $0.5857 \pm 0.0264$ & $0.5721 \pm 0.0158\blacktriangledown$ & $0.5365 \pm 0.0190\blacktriangledown$ & $0.5039 \pm 0.0215\blacktriangledown$ && $0.5819 \pm 0.0185$ & $0.5685 \pm 0.0177\blacktriangledown$ & $0.5458 \pm 0.0221\blacktriangledown$ & $0.5068 \pm 0.0175\blacktriangledown$ \\

\hline
\textbf{wins / loses} & 16/6 & 3/19 & 0/22 & 3/19 && 17/5 & 3/19 & 0/22 & 2/20 \\
\hline
\end{tabular}
\caption{\ourtechnique compared with Auto-Sklearn, TPOT and ML-Plan in terms of their balanced accuracy score (mean and standard deviation values are displayed)}
\label{tab:tool-comparison}
\end{sidewaystable}

This section addresses RQ2 by comparing \ourtechnique with Auto-Sklearn, TPOT, and ML-Plan. Table~\ref{tab:tool-comparison} presents the results of this comparison for budget durations of 1 and 6 hours. It displays the average balanced accuracy score over twenty repetitions, along with the standard deviation per dataset. To statistically validate our proposal, we employ a Wilcoxon signed-rank test with a significance level of 0.05. The p-values are adjusted for each budget and dataset pair using the Holm method. The symbols $\blacktriangle$ or $\blacktriangledown$ indicate whether the baseline approaches have significantly better or worse results than \ourtechnique. Finally, the table provides a summary of the number of wins and loses for each tool based solely on the average values.

As observed, \ourtechnique significantly outperforms Auto-Sklearn, TPOT, and ML-Plan across most datasets, irrespective of the time budget. For a 1-hour budget, \ourtechnique significantly outperforms Auto-Sklearn in fourteen out of twenty-two datasets, with Auto-Sklearn only surpassing it in the \textit{amazon} dataset. This result may be attributed to  \textit{amazon} having the highest number of features among datasets, excluding the sparse datasets \textit{dexter} and \textit{dorothea}. Notice that Auto-Sklearn uses BO, which aims to make as few evaluations as possible, and employs a warm-starting mechanism to start the optimisation process in promising regions of large search spaces. Similarly, \ourtechnique significantly outperforms TPOT and ML-Plan in fourteen and thirteen datasets, respectively, with neither of them outperforming \ourtechnique in any dataset. Additionally, it is noteworthy that \ourtechnique achieves significantly better performance than both TPOT and ML-Plan in eight out of ten multi-class datasets, indicating the suitability of our proposal for such problem domains. Similar trends are observed for a 6-hour budget, where \ourtechnique significantly outperforms Auto-Sklearn, TPOT, and ML-Plan in eleven, fourteen and fifteen datasets, respectively. Once again, none of the baseline approaches significantly outperforms \ourtechnique in any dataset.

Interestingly, there is little difference between using a 6-hour and a 1-hour budget. In fact, a larger budget can potentially harm results by increasing the risk of overfitting. For TPOT and \ourtechnique, a 6-hour budget outperforms a 1-hour budget in five and six datasets, respectively, with \textit{gisette}, \textit{amazon} and \textit{convex} being the common datasets. We also observe that datasets with more features, such as \textit{amazon}, benefit more from a larger budget than datasets with more samples. However, in two datasets (\textit{breastcancer} for TPOT and \textit{dexter} for \ourtechnique), the 1-hour budget outperforms the 6-hour budget. Specifically, the \textit{breastcancer} dataset (the smallest dataset) benefits more from a 1-hour budget, while \textit{dexter} (a sparse dataset) favors the 1-hour budget for \ourtechnique. For Auto-Sklearn, a 1-hour budget significantly improves results in seven datasets. However, in three datasets (\textit{amazon}, \textit{semeion}, and \textit{convex}), the 1-hour budget achieves significantly better results. Notably, \textit{amazon} and \textit{convex} are among the largest datasets in terms of features and samples, respectively. The warm-starting procedure of Auto-Sklearn could be a contributing factor. As mentioned above, Auto-Sklearn significantly outperforms \ourtechnique only in the \textit{amazon} dataset with a 1-hour budget. Increasing the budget of ML-Plan leads to significant improvements in results in four datasets without any performance decrease, which can be attributed to its mechanism for avoiding overfitting.

It is worth noting that there is no clear consensus on which datasets benefit from a larger budget. Specifically, there is no dataset for which results significantly improve with a 6-hour budget across all approaches. However, \textit{convex} and \textit{gisette} show improved results with a 6-hour budget for three of the approaches. Therefore, we increased the budget to 12 hours for both datasets. In this case, only Auto-Sklearn and \ourtechnique exhibit significant improvements in results, but only for the \textit{convex} dataset (approximately 1\% improvement in both cases). Similar to the 6-hour budget, \ourtechnique remains significantly superior to all baselines in both datasets.

\begin{figure}[H]
    \centering
     \begin{subfigure}[H]{0.49\textwidth}
        \centering
		 \includegraphics[width=1\textwidth]{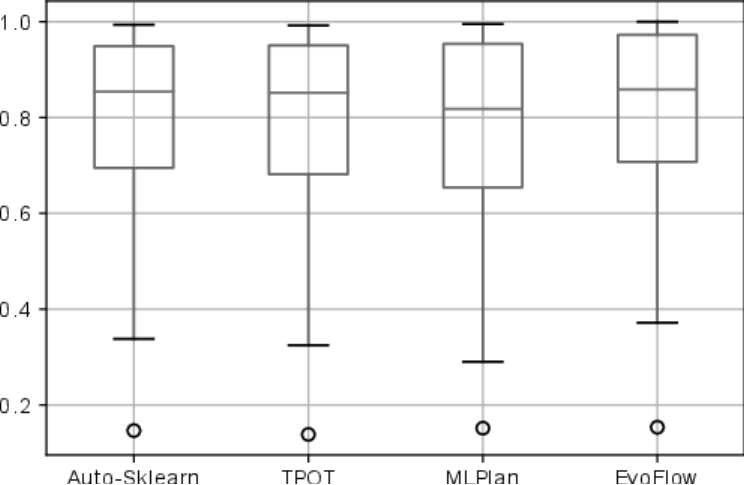}
        \caption{1-hour budget}
         \label{subfig:boxplot-1h}
    \end{subfigure}
    \hfill
    \begin{subfigure}[H]{0.49\textwidth}
        \centering
        \includegraphics[width=1\textwidth]{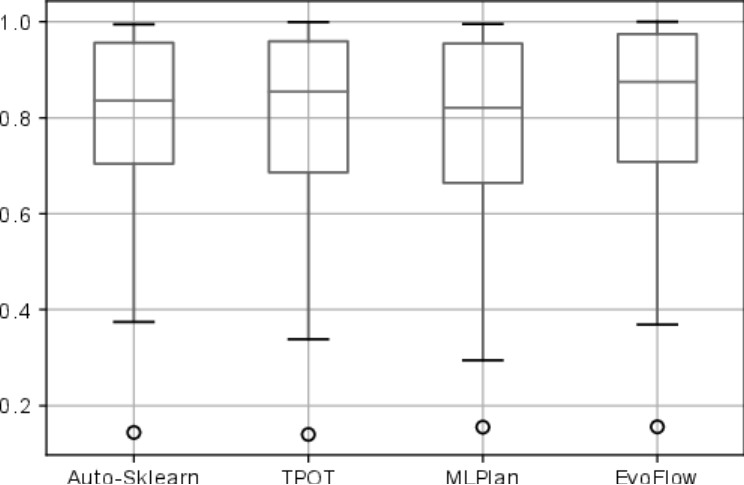}
        \caption{6-hour budget}
        \label{subfig:boxplot-6h}
    \end{subfigure}
    \caption{\ourtechnique compared to Auto-Sklearn, TPOT and ML-Plan}
    \label{fig:boxplot resume}
\end{figure}

\begin{figure}[H]
    \centering
     \begin{subfigure}[H]{0.6\textwidth}
        \centering
		 \includegraphics[width=1.1\textwidth]{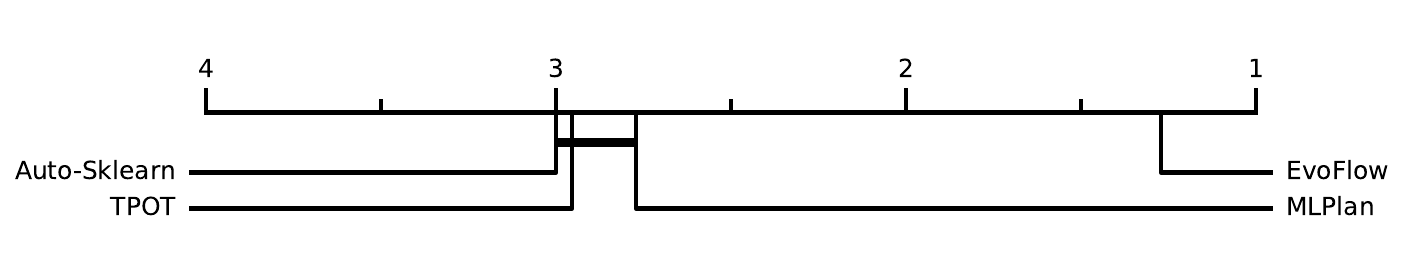}
        \caption{1-hour budget}
         \label{subfig:cd-1h}
    \end{subfigure}
    \\
    \begin{subfigure}[H]{0.6\textwidth}
        \centering
        \includegraphics[width=1.1\textwidth]{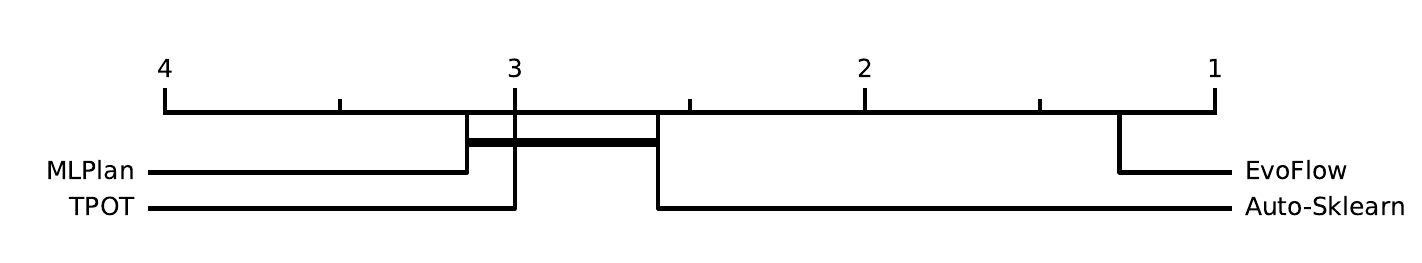}
        \caption{6-hour budget}
        \label{subfig:cd-6h}
    \end{subfigure}
    \caption{Critical difference diagram for \ourtechnique, Auto-Sklearn, TPOT and ML-Plan}
    \label{fig:cd-diagram}
\end{figure}

In summary, Figures~\ref{subfig:boxplot-1h} and \ref{subfig:boxplot-6h} display the distribution of balanced accuracy achieved by \ourtechnique and the comparison methods under 1-hour and 6-hour budgets, respectively. Here, \ourtechnique demonstrates the highest median performance in both scenarios. This finding aligns with the statistical analysis mentioned above, which is visually summarised in Figures~\ref{subfig:cd-1h} (1-hour budget) and ~\ref{subfig:cd-6h} (6-hour budget). These diagrams show that there are no significant differences among the three comparison methods (Auto-Sklearn, TPOT, and MLPlan).

\section{Experiment 3: \ourtechnique compared to RECIPE}
\label{sec:experiments_recipe}

\begin{table}[H]
\small
\centering
\begin{tabular}{llllll}
\hline
& \multicolumn{2}{c}{\textbf{1-hour budget}} && \multicolumn{2}{c}{\textbf{6-hour budget}} \\
\hline
 & \textbf{\ourtechnique} & \textbf{RECIPE} && \textbf{\ourtechnique} & \textbf{RECIPE} \\
\cline{2-3} \cline{5-6}
abalone & $0.1047 \pm 0.0054$ & - && $0.1082 \pm 0.0043$ & - \\
amazon & $0.8292 \pm 0.0137$ & - && $0.8400 \pm 0.0093$ & $0.7918 \pm 0.0385\blacktriangledown$ \\
breastcancer & $0.9843 \pm 0.0049$ & $0.9743 \pm 0.0073\blacktriangledown$ && $0.9840 \pm 0.0061$ & $0.9738 \pm 0.0090\blacktriangledown$ \\
car & $0.9991 \pm 0.0038$ & $0.9511 \pm 0.0217\blacktriangledown$ && $0.9977 \pm 0.0058$ & $0.9559 \pm 0.0163\blacktriangledown$ \\
convex & $0.7802 \pm 0.0141$ & - && $0.8028 \pm 0.0102$ & $0.7566 \pm 0.0144\blacktriangledown$ \\
dexter & $0.9530 \pm 0.0073$ & - && $0.9435 \pm 0.0099$ & $0.9434 \pm 0.0317$ \\
dorothea & $0.7538 \pm 0.0308$ & - && $0.7263 \pm 0.0261$ & - \\
germancredit & $0.6837 \pm 0.0146$ & $0.6142 \pm 0.0499\blacktriangledown$ && $0.6849 \pm 0.0147$ & $0.6292 \pm 0.0261\blacktriangledown$ \\
gisette & $0.9768 \pm 0.0020$ & - && $0.9793 \pm 0.0018$ & $0.9620 \pm 0.0530\blacktriangledown$ \\
glass & $0.7561 \pm 0.0252$ & $0.6441 \pm 0.0831\blacktriangledown$ && $0.7537 \pm 0.0431$ & $0.6730 \pm 0.0901\blacktriangledown$ \\
hillvalley & $1.0000 \pm 0.0000$ & $0.9468 \pm 0.0882\blacktriangledown$ && $1.0000 \pm 0.0000$ & $0.9501 \pm 0.0893\blacktriangledown$ \\
ionosphere & $0.9593 \pm 0.0175$ & $0.9389 \pm 0.0163\blacktriangledown$ && $0.9589 \pm 0.0186$ & $0.9435 \pm 0.0134\blacktriangledown$ \\
krvskp & $0.9945 \pm 0.0007$ & $0.9953 \pm 0.0016$ && $0.9947 \pm 0.0006$ & $0.9948 \pm 0.0025$ \\
madelon & $0.8563 \pm 0.0229$ & $0.8438 \pm 0.0312$ && $0.8894 \pm 0.0076$ & $0.8461 \pm 0.0324\blacktriangledown$ \\
secom & $0.5418 \pm 0.0386$ & - && $0.5489 \pm 0.0331$ & - \\
semeion & $0.9272 \pm 0.0097$ & $0.9181 \pm 0.0146\blacktriangledown$ && $0.9303 \pm 0.0125$ & $0.9181 \pm 0.0149\blacktriangledown$ \\
shuttle & $0.8982 \pm 0.0291$ & - && $0.8988 \pm 0.0384$ & $0.9452 \pm 0.1105\blacktriangle$ \\
spambase & $0.9557 \pm 0.0033$ & $0.9563 \pm 0.0046$ && $0.9566 \pm 0.0025$ & $0.9561 \pm 0.0045$ \\
waveform & $0.8597 \pm 0.0025$ & $0.8483 \pm 0.0102\blacktriangledown$ && $0.8596 \pm 0.0027$ & $0.8494 \pm 0.0104\blacktriangledown$ \\
winered & $0.3088 \pm 0.0180$ & $0.3225 \pm 0.0161$ && $0.3125 \pm 0.0124$ & $0.3235 \pm 0.0147\blacktriangle$ \\
winewhite & $0.3525 \pm 0.0188$ & $0.3936 \pm 0.0100\blacktriangle$ && $0.3568 \pm 0.0427$ & $0.3985 \pm 0.0090\blacktriangle$ \\
yeast & $0.5150 \pm 0.0307$ & $0.1099 \pm 0.0123\blacktriangledown$ && $0.5079 \pm 0.0254$ & $0.1077 \pm 0.0123\blacktriangledown$ \\

\hline
\textbf{wins/loses} & 10/14 & 4/14 && 15/19 & 4/19 \\
\hline
\end{tabular}
\caption{Comparison of \ourtechnique with G3P-based approach, RECIPE in terms of their $F_1$ score (mean and standard deviation values are displayed)}
\label{tab:recipe-comparison}
\end{table}

To address RQ3, we compare our approach with RECIPE, a pioneering technique applying G3P to tackle the AWC problem. Table~\ref{tab:recipe-comparison} presents the results of both approaches in terms of the $F_1$ score, which serves as the fitness function of RECIPE. It should be noted that for \ourtechnique, we use the solutions obtained from Section~\ref{subsec:experiments_comparison}, optimised based on the balanced accuracy score, and then compute their $F_1$ score for the purpose of comparison.

\begin{figure}[H]
    \centering
     \begin{subfigure}[H]{0.6\textwidth}
        \centering
		 \includegraphics[angle=90, width=1.1\textwidth]{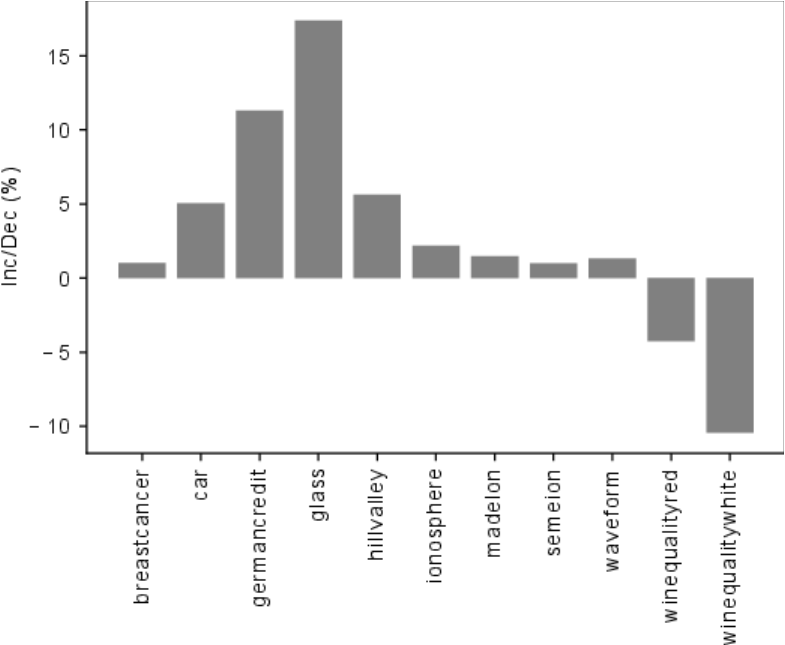}
        \caption{1-hour budget}
         \label{subfig:incdec-1h}
    \end{subfigure}
    \\
    \begin{subfigure}[H]{0.6\textwidth}
        \centering
        \includegraphics[angle=90, width=1.1\textwidth]{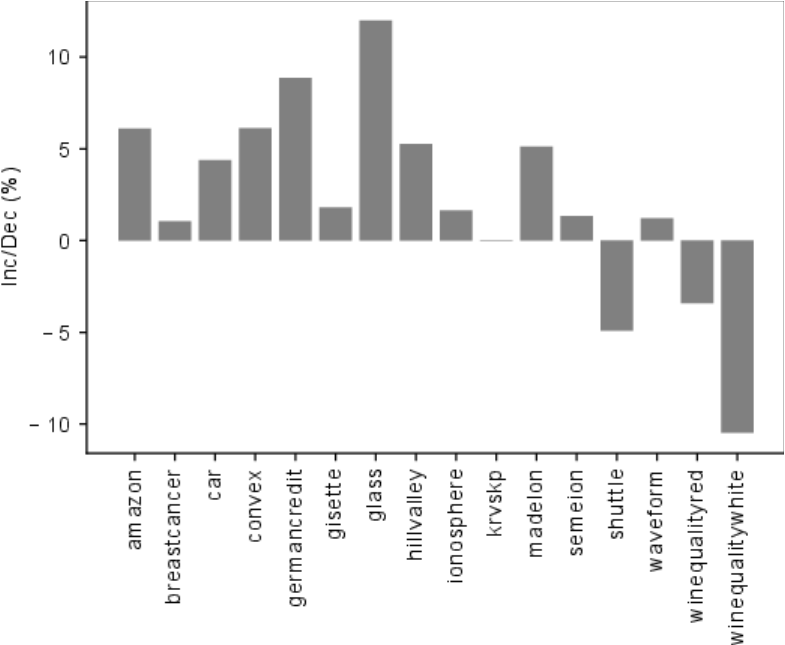}
        \caption{6-hour budget}
        \label{subfig:incdec-6h}
    \end{subfigure}
    \caption{Percentage of difference between \ourtechnique and RECIPE}
    \label{fig:incdec-diagram}
\end{figure}

As observed, RECIPE fails to produce results within the specified budgets for specific 
datasets (indicated as ``-''), and thus, these datasets are not considered in the ``wins/loses'' row. Despite this, \ourtechnique significantly outperforms RECIPE in nine and thirteen datasets for 1-hour and 6-hour budgets, respectively. Regardless of the budget, we observe that RECIPE does not always complete its optimisation process within the given time limits. This is likely due to RECIPE terminating the optimisation process after five generations without improving the best individual. Examining the 1-hour budget scenario (refer to Figure~\ref{subfig:incdec-1h}), the most notable performance differences are observed in the \textit{yeast}, \textit{glass}, and \textit{germancredit} datasets, where \ourtechnique surpasses RECIPE's results by 368\% (omitted for readability reasons), 17\%, and 11\%, respectively. It is important to note significant overfitting in the \textit{yeast} dataset, where RECIPE perfectly classifies all training samples in some runs. For the 6-hour budget scenario (Figure~\ref{subfig:incdec-6h}), the largest differences in favour of \ourtechnique are found in the same datasets, but the improvements are smaller. In this case, RECIPE shows improvements in the average values for the \textit{glass} and \textit{germancredit} datasets, while the results of \ourtechnique for these datasets do not improve. Conversely, RECIPE outperforms \ourtechnique in three datasets for both budgets: \textit{shuttle}, \textit{winewhite} (common to both budgets), and \textit{winered}. In fact, the most notable performance difference in favour of RECIPE is observed in the \textit{winewhite} dataset with an 11\% improvement. Finally, a Wilcoxon signed-rank test is conducted on the average values of each dataset for both budgets, confirming that \ourtechnique significantly outperforms RECIPE.

Regarding the budget comparison, it is worth mentioning that increasing the budget to 6 hours does not lead to significant improvements in the results of RECIPE for any dataset. This is likely due to RECIPE failing to produce any results within the 1-hour budget for the larger datasets, which would benefit the most from a budget increase. In fact, \ourtechnique exhibits significant improvements in its results with the budget increase for five datasets, including \textit{gisette}, \textit{amazon}, and \textit{convex}, for which RECIPE fails to generate any results within the 1-hour budget.

As a final note with a focus on the qualitative dimensions, it is worth highlighting that \ourtechnique produced superior solutions that could not be replicated by RECIPE due to its grammar specification. This phenomenon is exemplified in executions involving the \textit{germancredit} dataset. The grammar of RECIPE is designed to generate preprocessing sequences with a specific, fixed order and restricts the occurrence of specific 
operations to just one instance. These operations, though optional, encompass imputation, normalisation, scaling, feature selection, and feature generation. However, the best workflows, identified in some 
runs of \ourtechnique, incorporated more than one scaling algorithm, executed feature selection prior to scaling, or opted for imputation of missing values at later stages of the preprocessing sequence.

\section{Concluding remarks}
\label{sec:conclusions}

We have introduced \ourtechnique, a grammar-guided genetic programming algorithm designed to tackle the automated workflow composition problem. The use of a context-free grammar provides flexibility and customisability to our approach, enabling practitioners to adapt \ourtechnique according to their specific requirements. Unlike other evolutionary methods in the field, \ourtechnique incorporates genetic operators that are specifically designed to optimise workflows, encompassing both their structure and hyper-parameters. While it is common practice to construct ensembles from the best workflows discovered, we observed that as the evolution progresses, the population converges, resulting in similar workflows with identical predictions, \ie misclassifying the same samples. To mitigate this issue, \ourtechnique incorporates a mechanism for building ensembles that takes into account not only the predictive performance of the workflows but also the diversity of their predictions.

We have empirically validated \ourtechnique using a collection of classification datasets from the AWC literature. Initially, we compared different versions of \ourtechnique to establish that incorporating specific genetic operators and constructing ensembles of diverse workflows yields superior performance compared to the baselines. The results have demonstrated that combining these characteristics significantly outperforms the basic version of \ourtechnique, with the particular emphasis on the creation of diverse ensembles. Also, we pitted \ourtechnique against Auto-Sklearn, TPOT, ML-Plan and RECIPE, which use Bayesian optimisation, genetic programming, AI planning and grammar-guided genetic programming algorithms, respectively. The results have shown that \ourtechnique significantly outperforms in terms of predictive performance for the given time budgets in up to 68\% of the considered datasets, being statistically inferior in a marginal number of them.

In the future we plan to add more preprocessing and machine learning algorithms as grammar operators to support other learning tasks such as regression and clustering. Additionally, we believe it would be valuable to develop human-in-the-loop approaches to incorporate the expertise, experience, and intuition of data scientists into the optimisation process. Lastly, we intend to integrate \ourtechnique with popular tools like KNIME or RapidMiner to enhance accessibility and practicability for domain experts.

\section*{Supplementary material}
\label{sec:additional-material}
The source code of \ourtechnique is publicly available, along with a replication package containing all the necessary artefacts to reproduce the experiments outlined in this paper. The package includes the required scripts and information on the Python environments used. Information about the datasets and their partitions is available. Additionally, the raw results of the experiments conducted for both \ourtechnique and the baseline methods are provided. The complete statistical analysis, encompassing both unadjusted and adjusted p-values, is also reported. This supplementary material can be accessed from the following Zenodo repository:~\url{https://doi.org/10.5281/zenodo.10245033}

\section*{Acknowledgments}
\emph{Funding:} This work has been supported by the Spanish Ministry of Science, grant PID2020-115832GB-I00 funded by MICIN/AEI/10.13039/50110 0011033, and the Andalusian Regional Government (postdoctoral grant DOC\_00944).

\section*{Declaration of generative AI in scientific writing}
During the preparation of this work the authors eventually used ChatGPT 3.5 in order to improve readability and language of some paragraphs of the manuscript. This technology was always used with human oversight and control. After using this tool/service, the authors reviewed and edited the content as needed and take full responsibility for the content of the publication.

\bibliography{references}
\bibliographystyle{elsarticle-num} 

\end{document}